\documentclass[11pt]{article}

\usepackage[a4paper, margin=1in]{geometry}  
\usepackage[utf8]{inputenc} 
\usepackage[T1]{fontenc} 
\usepackage{titletoc}
\usepackage{hyperref}       
\usepackage{url}            
\usepackage{booktabs}       
\usepackage{amsfonts}       
\usepackage{nicefrac}       
\usepackage{microtype}      
\usepackage{xcolor}         
\usepackage{graphicx}       
\usepackage[ruled,vlined]{algorithm2e}  
\usepackage{amssymb, amsthm, amsmath, xfrac, cases}
\usepackage{mathtools}
\usepackage{hyperref}    
\usepackage{lmodern} 

\hypersetup{colorlinks, linkcolor={red!70!black}, citecolor={blue!70!black},
urlcolor={red!70!black} }

\theoremstyle{plain}
\newtheorem{theorem}{Theorem}[section]

\newtheorem{result}[theorem]{Result}

\theoremstyle{definition}

\theoremstyle{remark}

\usepackage{enumitem}

\usepackage{natbib}
\setcitestyle{authoryear,round,citesep={;},aysep={,},yysep={;}}
\usepackage{custom} 
\renewcommand{\vec}[1]{\boldsymbol{#1}}

  \makeatletter
  \def\sign{\@ifnextchar*{\@sgnargscaled}{\@ifnextchar[{\sgnargscaleas}{\@ifnextchar{\bgroup}{\@sgnarg}{\sgn} }}}
  \def\@sgnarg#1{\sgn\rbr{#1}}
  \def\@sgnargscaled#1{\sgn\rbr*{#1}}
  \def\@sgnargscaleas[#1]#2{\sgn\rbr[#1]{#2}}
  \makeatother


\newcommand{\speedup}[1]{{\color{gray}(\ifdim #1 pt > 0.3pt #1\else $< #1$\fi{}$\times$)}}
\usepackage{amsfonts}
\usepackage{amsmath}
\usepackage{amsthm}
\usepackage{amssymb}
\usepackage{dsfont}

\usepackage{xcolor}
\usepackage{color}
\usepackage{graphicx}

\usepackage{verbatim}
\usepackage{xspace} %

\usepackage{enumerate}
\usepackage{enumitem}

\makeatletter
\newsavebox{\@brx}
\newcommand{\llangle}[1][]{\savebox{\@brx}{\(\m@th{#1\langle}\)}%
  \mathopen{\copy\@brx\mkern2mu\kern-0.9\wd\@brx\usebox{\@brx}}}
\newcommand{\rrangle}[1][]{\savebox{\@brx}{\(\m@th{#1\rangle}\)}%
  \mathclose{\copy\@brx\mkern2mu\kern-0.9\wd\@brx\usebox{\@brx}}}
\makeatother

\renewcommand{\vec}{\mathbf}
  \usepackage{bm}

\providecommand{\mycomment}[3]{\todo[caption={},size=footnotesize,color=#1!20]{\textbf{#2: }#3}}%
\providecommand{\inlinecomment}[3]{%
  {\color{#1}#2: #3}}%
\newcommand\commenter[2]%
{%
  \expandafter\newcommand\csname i#1\endcsname[1]{\inlinecomment{#2}{#1}{##1}}
  \expandafter\newcommand\csname #1\endcsname[1]{\mycomment{#2}{#1}{##1}}
}

  \definecolor{mydarkblue}{rgb}{0,0.08,0.45}
  \usepackage{hyperref}
  \hypersetup{ %
    pdftitle={},
    pdfauthor={},
    pdfsubject={},
    pdfkeywords={},
    pdfborder=0 0 0,
    pdfpagemode=UseNone,
    colorlinks=true,
    linkcolor=mydarkblue,
    citecolor=mydarkblue,
    filecolor=mydarkblue,
    urlcolor=mydarkblue,
    pdfview=FitH}

\usepackage{url}

\hypersetup{citecolor=blue}
\hypersetup{
    citecolor=blue,
    linkcolor=blue,
    urlcolor=red!70!black
}

\title{Attention Head Specialization:\\ Phase Transitions in Sample Complexity}

\author{
Léo Catteau$^1$,
Odilon Duranthon$^1$,
Lenka Zdeborov\'a$^1$
}
\date{
\small
$^1$ Statistical Physics of Computation Laboratory,
\\
École Polytechnique Fédérale de Lausanne (EPFL)
CH-1015 Lausanne
}

\usepackage{xcolor}
\definecolor{PS}{HTML}{989898}
\definecolor{MEM}{HTML}{D67853}
\definecolor{SPEC}{HTML}{64A8C3}
\definecolor{2-SPEC}{HTML}{397993}

\begin{document}
\maketitle

\begin{abstract}
    Attention heads specialize into distinct functions during training, often emerging in stages together with sharp changes in model behavior. While this phenomenon has mostly been studied as a function of training time, we ask how head specialization depends on the amount of available training data. Empirically, we find that heads specialized to syntactic relations in BERT emerge as the dataset size increases. We then study this phenomenon in a solvable high-dimensional model of multi-head attention and establish that, in this model, head specialization is a genuine phase transition in sample complexity, corresponding to a non-analyticity in the test error and other observables.  Depending on data structure and regularization, specialization can follow uninformative head differentiation or occur through a cascade of phase transitions that progressively recover weaker latent features, providing a statistical form of simplicity bias. Finally, we relate these statistical transitions to the stagewise emergence of specialized heads during online learning, disentangling statistical from dynamical origins of head specialization.
\end{abstract}

\section{Introduction}
Multi-head attention is a central architectural ingredient of modern transformer models, enabling multiple attention patterns to coexist within a single layer. Different heads learn to perform distinct functions, while their collaboration allows implementing circuits that no individual head captures alone. Empirical studies have identified such specialized heads across multiple settings, including induction heads involved in in-context learning (ICL) \citep{olsson2022context,pmlr-v267-yin25e},  heads specializing to different syntactic relations \citep{ICLR2024_3e8df255}, and heads progressively differentiating into distinct functional roles during training \citep{wang2025differentiation}. Head specialization has predominantly been studied as a function of training time: specialized mechanisms emerge at distinct stages of training, sometimes coinciding with sharp improvements in performance or drops in training loss \citep{hoogland2024loss,aoyama2026predicting}.

\begin{figure}[tbp]
    \centering
    \includegraphics[width=1\linewidth]{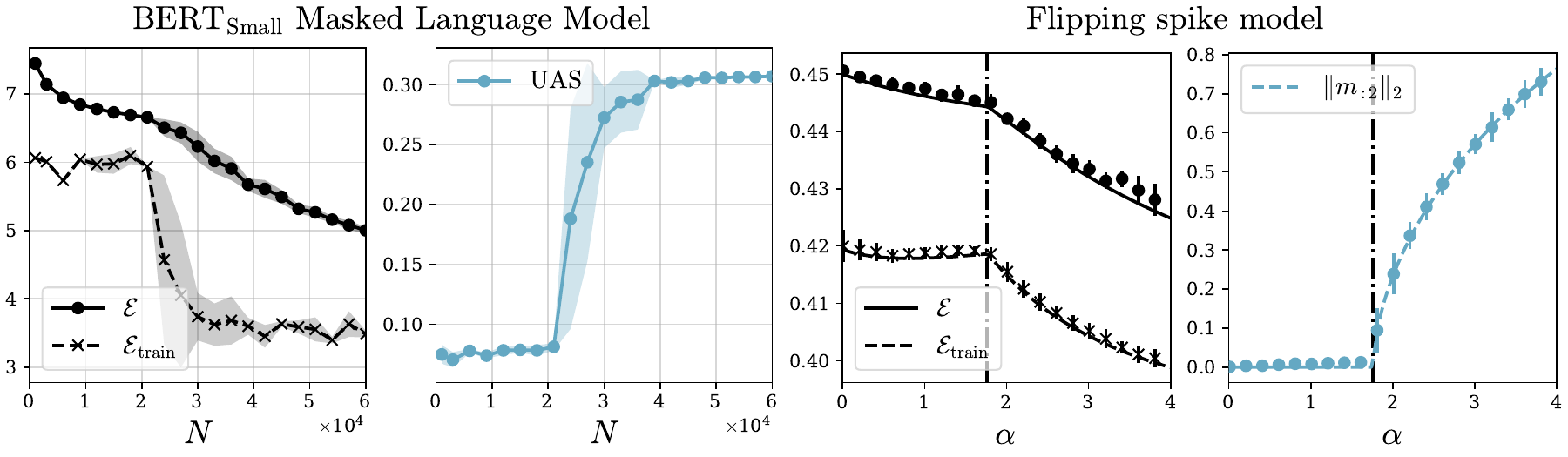}
    \caption{\textbf{Head specialization as a function of sample complexity.} \textbf{Left:} $\mathrm{BERT}_\mathrm{Small}$ models trained independently on fixed datasets of $N$ sequences through minimization of masked-language-modeling (MLM) loss. \textbf{(first panel)} Validation error and training MLM loss (\ref{eq:MLM_loss}). \textbf{(second panel)} Unlabeled attachment score (UAS)(\ref{eq:BERT_UAS}), measuring the fraction of dependency parents correctly recovered from word-level attention by the best head, layer, and attention direction. Curves show averages over five runs, with shaded bands indicating seed variability. Reported metrics are evaluated at early-stopped checkpoints selected by the lowest validation loss along each training trajectory. Experimental details are provided in App.~\ref{App:BERT}. \textbf{Right:} Qualitative comparison with the solvable flipping spike model (Sec.~\ref{model}) as the sample complexity $\alpha=N/D$ increases, with $N$ the number of sequences and $D$ the token dimension. Solid lines correspond to theoretical predictions and markers to numerical experiments with $D=2000$, $\nu_1=0.35$, $\nu_2=2.0$, $L=10$, $H=2$, $\lambda=0.2$. \textbf{(third panel)} Test and training errors exhibit non-analyticity at the specialization threshold. \textbf{(fourth panel)} Head--feature alignment reveals specialization to the latent feature.}
    \label{Fig:0}
\end{figure}

Much less is known about specialization as a function of the amount of available training data. Rather than asking \emph{when} during training different heads specialize, one can ask \emph{how many samples} are needed for distinct specialized heads to emerge. This distinction separates a statistical question about the information available in the dataset from the particular optimization dynamics used to learn it. We investigate this question in BERT masked language models (MLMs), where syntactic attention structure (SAS), characterized by heads specializing to different syntactic relations \citep{clark2019does}, emerges during training alongside a sharp drop in loss \citep{ICLR2024_3e8df255}. Empirically, training $\mathrm{BERT}_\mathrm{Small}$ models from random initialization on datasets of $N$ sequences from BookCorpus and English Wikipedia reveals a clear phenomenology shown in Fig.~\ref{Fig:0}: as $N$ increases, SAS emerges sharply over a characteristic range of dataset sizes, accompanied by abrupt changes in the learning curves. For each dataset size, we fix the sequences and masking patterns throughout training and report metrics at the checkpoint with lowest validation MLM loss. Following \citep{ICLR2024_3e8df255}, we quantify SAS on a separate syntactically annotated corpus as the fraction of parent-child pairs recovered by each head's maximum word-level attention. Aggregating accuracy across relations for the best head gives the unlabeled attachment score (UAS), which measures alignment between attention and syntactic structure enabled by specialization of the heads. 

Sample complexity provides a natural control parameter for asking whether such emergence corresponds to a genuine phase transition. In the statistical physics of learning, one considers a proportional high-dimensional limit in which the number of trainable degrees of freedom and training samples grow together, $D,N\to\infty$ at fixed $\alpha=N/D$. This framework has a long history in neural-network theory, where increasing sample complexity can induce phase transitions from permutation symmetric to specialized solutions in the committee machine \citep{schwarze1992generalization,schwarze1993generalization,aubin2018committee}. In this limit, changes between qualitatively different learning regimes can become sharp, with nonanalyticities of macroscopic observables at critical values of~$\alpha$. For multi-head attention, this raises a precise question: can the emergence of specialized heads be understood as a genuine phase transition controlled by sample complexity?

We answer this question in a tractable multi-feature variant of the single-location regression (SLR) model \citep{marion2025attention,sagitova2026specializationsoftmaxattentionheads}, whose head-specialization dynamics under online learning were characterized in \citet{sagitova2026specializationsoftmaxattentionheads}. The task is to retrieve one token carrying a low-dimensional latent signal from a sequence of otherwise uninformative tokens, using multiple attention heads that can specialize to different latent features. Here, instead of following training dynamics, we study the associated empirical risk minimization (ERM) problem as a function of sample complexity. Fig.~\ref{Fig:0} shows that this model exhibits behavior qualitatively analogous to that observed in the $\mathrm{BERT}_\mathrm{Small}$ experiment. Moreover, we show that, in this model, head specialization occurs through phase transitions in sample complexity, characterized by non-analyticities in the order parameters and prediction error at critical values of~$\alpha$.

The resulting phase diagram reveals a richer structure than a single specialization threshold. At low regularization and low sample complexity, specialization is preceded by an unspecialized phase with broken head-permutation symmetry, in which heads differ by fitting sample-specific fluctuations without recovering distinct latent features. This illustrates that distinct attention patterns alone do not imply specialization that improves generalization. For heterogeneous signals, increasing sample complexity instead produces a cascade of specialization transitions: heads successively recover latent features from strongest to weakest, so that increasingly subtle structure requires increasingly more data. This provides a statistical counterpart of stagewise learning and simplicity bias, arising here from the structure of the ERM solutions rather than from a particular training dynamics.

Finally, we connect these equilibrium transitions back to the training-time phenomenology from which we started \citep{olsson2022context,wang2025differentiation,aoyama2026predicting, ICLR2024_3e8df255}. In online learning, training time and the amount of observed data increase together, entangling the statistical transition with optimization dynamics. In the solvable model, we show how the critical sample complexities map onto the emergence of specialized heads along training. This perspective suggests that some of the abrupt or stagewise emergence observed during transformer training may reflect underlying statistical phase transitions, while distinguishing these from phenomena that are genuinely dynamical and helping clarify how such transitions manifest in learning dynamics.

\paragraph{Main Contributions}
The main empirical result of the paper is the characterization of head specialization in the number of training samples. In a tractable model of multi-head attention, we then show that head specialization is a statistical phase transition in sample complexity, and connect the equilibrium transition to specialization during training. Our main technical contributions are:
\begin{itemize}
    \item We derive a tight asymptotic characterization of the regularized ERM problem in the proportional high-dimensional limit $N,D\to\infty$ at fixed $\alpha=N/D$, through closed self-consistent equations for the errors and order parameters (Sec.\ref{Sec:asymptotic_characterization}).
    \item We show that head specialization is a genuine phase transition in sample complexity. Depending on regularization and signal structure, we uncover an unspecialized phase with broken head-permutation symmetry, as well as a cascade of specialization transitions in which increasingly weaker latent features are recovered with increasing sample complexity, yielding a statistical form of simplicity bias (Sec.\ref{specialization}).
    \item We establish a controlled correspondence between these equilibrium phase transitions and the emergence of specialized heads during online learning, relating critical sample complexities to training-time specialization and disentangling statistical from optimization-dependent effects (Sec.\ref{dynamics}).
\end{itemize}

\paragraph{Notations.}
We write $[n]=\{1,\ldots,n\}$, $\delta_{ij}$ for the Kronecker delta, and $\mathbb{I}_n$ for the identity matrix. For a matrix $A$, $A_i$ and $A_{:j}$ denote its $i$-th row and $j$-th column. We write $\lVert\cdot\rVert_2$ for the Euclidean norm, $\lVert\cdot\rVert_F$ for the Frobenius norm, and $v^{\otimes 2}=vv^\top$. We denote by $\mathcal{N}(\mu,\Sigma)$ the Gaussian distribution with mean $\mu$ and covariance $\Sigma$.

\subsection*{Related works}
\paragraph{Empirical emergence and head specialization.}
Specialized attention heads have been observed across architectures and tasks, including induction heads associated with in-context learning \citep{olsson2022context,pmlr-v267-yin25e}, syntactic heads in masked language models \citep{ICLR2024_3e8df255}, and circuits studied in controlled synthetic settings \citep{pmlr-v235-singh24c}. A complementary literature studies when such capabilities and internal mechanisms emerge during training, including abrupt changes in performance \citep{wei2022emergent}, stagewise development and head differentiation \citep{hoogland2024loss,wang2025differentiation}, and the emergence time of induction heads \citep{aoyama2026predicting}. Such behavior can depend strongly on data, initialization, and training trajectory: emergent capabilities can be transient \citep{NEURIPS2023_58692a17}, vary across random seeds \citep{juneja2023linear,zhao2026random}, and depend on the structure of the training data \citep{qin-etal-2025-data}. While these works predominantly follow a model through training, our BERT experiments instead hold dataset size fixed for each training run, and study how syntactic head specialization changes across datasets of different sizes.

\paragraph{Theory of attention and head specialization.}
Recent solvable models characterize learning in attention architectures, including single-location regression \citep{marion2025attention,ICLR2026_8c21b98a}, sequence multi-index models \citep{troiani2025fundamental}, attention-indexed models \citep{boncoraglio2025single}, mapping to generalized Potts model \citep{rende2024mapping}, or hierarchical models \citep{cagnetta2024towards}. Multi-head specialization has primarily been studied through training dynamics \citep{chen2024training,NEURIPS2023_58692a17,saxe25iclLinear,sagitova2026specializationsoftmaxattentionheads,yüksel2026incrementallearningsparseattention}. Here we instead study specialization at equilibrium as a function of sample complexity.

\paragraph{Phase transitions and specialization in neural networks.}
Specialization as a sample-complexity phase transition has a long history in committee machines, where increasing $\alpha=N/D$ can induce transitions to solutions in which different hidden units align with distinct features of the data \citep{schwarze1992generalization,schwarze1993generalization,aubin2018committee,ortiz26committee}. Phase transitions have also been studied in solvable attention models \citep{NEURIPS2024_3fefebc2}; most closely, \citet{lavie2026phasetransitionsattentionbayesian} establish a sample-complexity transition for the emergence of a single copy head. We instead study specialization of attention heads and the associated emerging generalization capabilities.

\section{Single-location regression model}\label{model}
We consider the multi-feature spiked single-location regression (SLR) model of \citet{sagitova2026specializationsoftmaxattentionheads}. The task is to retrieve one randomly located relevant token from a sequence of otherwise uninformative tokens. The informative token is structured by a low rank planted signal over $F$ latent directions. Learning this structure with multiple low-rank heads provides a tractable setting for studying head specialization.

\paragraph{Data model.}
We consider an embedding dimension $D$. We first draw $F$ independent hidden spikes $k_f^*\sim \mathcal{N}(0, \mathbb{I}_D)$ defining the latent structure that remains fixed across all sequences.
We denote $K^*=\{k_f^*\}_{f=1}^F$ the matrix with rows $k_f^{*\top}$. A sequence $X\in\mathbb{R}^{L\times D}$ consists of $L$ tokens $\{X_{\ell}\}_{1\leq \ell\leq L}$. A hidden index $\epsilon\in [L]$ identifies the relevant token $X_\epsilon$ containing a planted spike $\hat{k}$ generated from the $F$ latent directions, while remaining tokens $X_{\ell \neq \epsilon}$ are independent isotropic Gaussian noise. For each sequence, we independently draw $\epsilon \sim \text{Unif}([L])$ and the feature weights $\theta\sim P_\theta$ with $\theta \in \mathbb{R}^F$,  defining the sequence-dependent spike $\hat k$, such that the tokens $X_\ell$ are independently distributed conditionally on $\epsilon$ and $\theta$ as
\begin{align}
    \hat k = \frac{K^{*\top}\theta}{\sqrt{D}} \in \mathbb{R}^D, 
    \quad  \quad   \quad 
    X_\ell \sim \mathcal{N}(\delta_{\ell,\epsilon}\hat k , \mathbb{I}_D), \quad \ell \in [L].
\end{align}
The associated label is the relevant token $y = X_\epsilon$. As the latent $\epsilon$ and $\theta$ are sampled at every sequence/context, deducing them can also be seen as a simple case of in-context learning. In the following, we will consider two choices for the distribution $P_\theta$. In the $\textbf{flipping spike model}$, a latent direction is kept constant while another flips sign randomly across sequences:  $F=2$, $\theta_1 = \sqrt{\nu_1}$, and $\theta_2\sim \mathrm{Unif}(\{-\sqrt{\nu_2},\sqrt{\nu_2}\})$, where we define signal strengths $\nu_1>0$, $\nu_2>0$. In the $\textbf{non-isotropic Gaussian model}$, all components of the planted spike fluctuate independently with different covariances: $\theta_f\sim \mathcal{N}(0, \nu_f)$, where we define signal strengths $\nu_1 \geq ... \geq \nu_F\geq 0$. 

\paragraph{Multi-head attention.}  We consider learning of the stated task with a single layer of low-rank attention with $H$ heads. This architecture is well suited for retrieval of low rank structure, with different heads attending to different features. Specifically, we consider the estimator $f_{K}:\mathbb{R}^{L\times D}\to \mathbb{R}^D$ given by rank-one attention heads with trainable weights $K=\{k_h\}_{h=1}^H\in \mathbb{R}^{H\times D}$:
\begin{align}\label{eq:estimator}
    f_K(X) = \frac{1}{H}\sum_{h=1}^H \text{softmax}\left(\frac{Xk_h}{\sqrt{D}} \right)^\top X,
\end{align}
where attention vectors are uniformly averaged across heads. This model can be viewed as a simplification of cross-attention with fixed sequence-independent query embedding and identity value matrix, as discussed in \citet{marion2025attention} and \citet{ICLR2026_8c21b98a}. Note that, while considering larger (extensive) rank attention is technically possible \citep{boncoraglio2025single}, this is left for future work. Similarly, training the value matrix with a fixed number of parameters could be accommodated by an extended parametrization. The current model then constitutes a relevant minimal estimator for the study of head specialization in the designed task. As mentioned in \citet{sagitova2026specializationsoftmaxattentionheads}, this type of attention layer is also used in practice \citep{carion2020endtoendobjectdetectiontransformers}.

\paragraph{Empirical risk minimization.} We study learning from a finite dataset $\mathcal{D}= \{X_\mu, y_\mu\}_{\mu = 1}^N$ of $N$ independently sampled sequences, and their associated labels. An attention layer $f_{\hat K}$ is learned by regularized empirical risk minimization (ERM):
\begin{align}\label{eq:ERM}
    \hat K = \underset{K\in \mathbb{R}^{H\times D}}{\text{argmin}} {\mathcal{L}}_{\rm train}(K), \quad \quad  {\cal L}_{\rm train}(K) = \sum_{\mu=1}^N \frac{1}{2D} \lVert y_\mu - f_K(X_\mu) \rVert_2^2 + \frac{\lambda}{2}\lVert K \rVert_F^2
\end{align}
where $\lambda>0$ is the $\ell_2$ regularization strength. The performance of the estimator (\ref{eq:ERM}) is evaluated through the training error $\mathcal{E}_\mathrm{train}$ associated with the objective (\ref{eq:ERM}), and the test error $\mathcal{E}$:
\begin{align}\label{eq:test_error}
    \mathcal{E}_\mathrm{train}=\frac{1}{N}\sum_{\mu=1}^N \frac{1}{2D} \lVert y_\mu - f_{\hat K}(X_\mu) \rVert_2^2,  &&\mathcal{E} = \mathbb{E}_{X, y}\left[ \frac{1}{2D} \lVert y- f_{\hat K}(X)\rVert_2^2\right].
\end{align}

\paragraph{High-dimensional limit.}
We analyze the ERM problem (\ref{eq:ERM}) in the proportional limit where $D,N\to \infty$ while the ratio $\alpha=N/D$, the sample complexity, remains of order $\Theta(1)$. In order to study head specialization, we further assume the number of heads $H$, the number of features $F$, the sequence length $L$, and the regularization strength $\lambda$ to be $\Theta(1)$. This proportional limit has been considered in a stream of previous works (e.g. \citet{NEURIPS2024_3fefebc2,troiani2025fundamental,Cui_2024,ICLR2026_8c21b98a}), and allows deriving a closed-form characterization of the ERM problem which we present in the next section. As the number of trainable degrees of freedom grows linearly with $D$, taking $N=\Theta(D)$ retains nontrivial interactions between the dataset and the architecture, and sample complexity $\alpha$ appears as a natural control parameter to study head-specialization phase transitions. This particular scaling is, however, setting-specific: other well-defined control parameters may arise in different high-dimensional limits (e.g. extensive-rank architectures \citet{pmlr-v119-gerace20a, NEURIPS2025_8039ca1e, boncoraglio2025single, barbier2026statistical}), or be defined through properties of the data.

\section{Asymptotic characterization of the empirical risk}\label{Sec:asymptotic_characterization}
The main technical result of this work is a closed-form characterization of the minimizers of the regularized empirical risk (\ref{eq:ERM}) obtained in the high-dimensional limit for the model (\ref{eq:estimator}) through the test error (\ref{eq:test_error}), training error (\ref{eq:test_error}) and additional summary statistics as functions of sample complexity. Defined from the high-dimensional weights, these summary statistics concentrate in the high-dimensional limit and fully describe the model's macroscopic state: the head--feature overlaps $m_{hf}$ quantify the alignment between each head's learned attention vector $k_h$ and each planted feature direction $k_f^*$, while the self-overlaps $q_{hh'}$ quantify the alignment between learned attention vectors $k_h$ and $k_{h'}$, with diagonal entries encoding their normalized squared norms.
\begin{result}\label{res:self-consistent_equations}
Under the assumption of replica symmetry, in the limit
$N,D\rightarrow\infty$, with
$H,\allowbreak F,\allowbreak L,\allowbreak
\lVert\theta\rVert_2,\allowbreak \lambda,\allowbreak
N/D=\Theta_D(1)$, the summary statistics
\begin{align}\label{eq:summary_statistics}
	m= \frac{K K^{*\top}}{D} \in \mathbb{R}^{H\times F}, && q = \frac{K K^\top}{D}\in\mathbb{R}^{H\times H}
\end{align} 
concentrate in probability, and are solutions of the set of finite-dimensional self-consistent equations
\begin{align}\label{eq:saddle_point_equations}&
    \begin{cases}
        m=(\lambda \mathbb{I}_H+\hat V)^{-1}\hat m\\ 
        q=(\lambda \mathbb{I}_H+\hat V)^{-1}(\hat m \hat m^\top+\hat q)(\lambda\mathbb{I}_H+\hat V)^{-1}\\
        V = (\lambda\mathbb{I}_H+\hat V)^{-1}
    \end{cases}\\&
    \begin{cases}\label{eq:saddle_point_equations_end}
        \hat m=\alpha \mathbb{E}_{\theta, \Xi}\left[V^{-1}\left( \mathrm{prox}(\theta, \Xi)_1 - \omega_1\right)\theta^\top\right]\\
        \hat q=\alpha \mathbb{E}_{\theta, \Xi}\left[V^{-1}\left(\sum_{\ell=1}^L\left( \mathrm{prox}(\theta, \Xi)_\ell - \omega_\ell\right)^{\otimes 2} \right)V^{-1}\right]\\
        \hat V=-\alpha \mathbb{E}_{\theta, \Xi}\left[V^{-1}\left(\sum_{\ell=1}^L\left(\mathrm{prox}(\theta, \Xi)_\ell - \omega_\ell \right)\xi_\ell^\top \right)q^{-1/2} \right],
    \end{cases}
\end{align}
with the summary statistics $\{m,q,V\}$ and their conjugates $\{\hat m,\hat q,\hat V\}$. In \textup{(\ref{eq:saddle_point_equations}-\ref{eq:saddle_point_equations_end})}, we define the Gaussian random variables $\Xi=\{\xi_\ell\}_{\ell=1}^L$ with $\xi_\ell\sim\mathcal{N}(0,\mathbb{I}_H)$ and the associated local fields $\Omega = \{\omega_\ell\}_{\ell=1}^L$ with $\omega_\ell \coloneq \delta_{\ell,1}m\theta + q^{1/2}\xi_\ell$. The proximals $\{\mathrm{prox}(\theta, \Xi)_\ell\}_{\ell=1}^L$ are defined as minimizers of the Moreau envelope 
{\small
\begin{align}
    \mathcal{M}(\theta, \Xi)=\inf_{Z\in\mathbb{R}^{L\times H}}\left\{ \frac{1}{2}\sum_{\ell=1}^L\left(\delta_{\ell1} - \frac{1}{H}\sum_{h=1}^H \mathrm{softmax}(Z_{:h})_\ell\right)^2 + \frac{1}{2}\sum_{\ell=1}^L \left(Z_\ell-\omega_\ell\right)^\top V^{-1}\left(Z_\ell-\omega_\ell\right)\right\}.
\end{align}}
In the same limit, the test error \eqref{eq:test_error} converges in probability to 
\begin{align}\label{eq:macroscopic_test_error}
    \mathcal{E} =\frac{1}{2}\mathbb{E}_{\theta,\Xi}\left[\sum_{\ell=1}^L\left( \delta_{\ell 1}- \frac{1}{H}\sum_{h=1}^H\mathrm{softmax}(\Omega_{:h})_\ell\right)^2 \right].
\end{align}
Finally, the training error at the ERM minimizer converges in probability to 
\begin{align}\label{eq:macroscopic_training_loss}
    \mathcal{E}_\mathrm{train} =  \mathbb{E}_{\theta, \Xi}\left[ \mathcal{M}\left( \theta, \Xi\right)\right] - \frac{1}{2\alpha}\mathrm{Tr}\left( \hat q V\right) .
\end{align}
\end{result}

Result \ref{res:self-consistent_equations} provides a tight asymptotic characterization of the test error $\mathcal{E}$ (\ref{eq:macroscopic_test_error}) and the training error $\mathcal{E}_\mathrm{train}$ (\ref{eq:macroscopic_training_loss}) as a function of finite-dimensional summary statistics $\{m,q,V\}$ characterized in closed form by the set of equations (\ref{eq:saddle_point_equations}-\ref{eq:saddle_point_equations_end}), providing a finite-dimensional description of the high dimensional learning problem (\ref{eq:ERM}). Solutions of the self-consistent equations (\ref{eq:saddle_point_equations}-\ref{eq:saddle_point_equations_end}) asymptotically describes the global minimum of (\ref{eq:ERM}) as well as saddles and local minima. Multiple solutions can coexist and describe competing minima of the training loss. Thus, changing the control parameters such as sample complexity can change the stability of minima or selection of the global minimum, and produce phase transitions.

The derivation of Result~\ref{res:self-consistent_equations}, detailed in App.~\ref{app:replica_derivation}, relies on the replica method under a replica-symmetric (RS) ansatz. While replica-symmetric predictions have been rigorously established for convex ERM problems (see, e.g., \cite{aubin2020generalization}) and under specific conditions for non-convex ones \citep{vilucchio2025asymptotics}, their validity in our non-convex setting remains an assumption. The resulting characterization is supported by numerical experiments: in the regimes studied, training the attention layer (\ref{eq:estimator}) on the ERM objective (\ref{eq:ERM}) by gradient-based optimization from random initializations (App.~\ref{App:Experiments}) yields solutions whose summary statistics, training error and test error agree with those of the predicted global minima.  Finally, we mention that while Result \ref{res:self-consistent_equations} is presented specifically with softmax activation of the training loss for clarity, similar results can be reached for generic activation function. 

\section{Head-specialization phase transition}\label{specialization}
\paragraph{Phases and phase transitions.} We characterize the phase transitions that emerge in the high-dimensional limit, through the non-analyticities of the order parameters and the test error in the sample complexity. The empirical risk (\ref{eq:ERM}) is invariant under permutation of the heads, while the self-consistent equations (\ref{eq:saddle_point_equations}-\ref{eq:saddle_point_equations_end}) additionally inherit the symmetries of the data distribution. We call a phase specialized when the heads have distinct overlaps with the planted features, and unspecialized otherwise. At finite sample complexity, however, head weights can grow and correlate while being orthogonal to the planted features, allowing for the existence of an unspecialized phase breaking head-permutation symmetry in the self-overlap $q$.

For the flipping-spike model with $H=2$, $F=2$, the data distribution is invariant under sign reversal of the second feature. We distinguish the permutation symmetric (PS), permutation-symmetry-broken (PSB), and specialized (SPEC) phases using the head-feature overlap $m$ and antisymmetric self-overlap $q_\perp \equiv (q_{11}+q_{22}-2q_{12})/2$,
which measures the squared distance between the attention-head vectors. The three phases satisfy
\begin{align}
    \text{\textcolor{PS}{PS}}:\quad&
    m_{11}=m_{21}>0,\quad m_{12}=m_{22}=0,\quad q_\perp=0,
    \\
    \text{\textcolor{MEM}{PSB}}:\quad&
    m_{11}=m_{21}>0,\quad m_{12}=m_{22}=0,\quad q_\perp>0,
    \\
    \text{\textcolor{SPEC}{SPEC}}:\quad&
    m_{11}=m_{21}>0,\quad m_{12}=-m_{22}\neq0,\quad q_\perp>0.
\end{align} 
Thus, \textcolor{MEM}{PSB} distinguishes the heads without aligning their difference with the flipping feature, whereas \textcolor{SPEC}{SPEC} develops opposite nonzero overlaps with this feature.

For the non-isotropic Gaussian model with distinct feature strengths ordered in decreasing order, the solutions studied below exhibit successive specialized phases ($r$-SPEC) indexed by $r\in\{0,\ldots,\min(F,\lfloor H/2\rfloor)\}$ and characterized by
\begin{align}
    \text{\textcolor{SPEC}{$r$-SPEC}:}\quad
    \begin{cases}
        \lVert m_{:f}\rVert_2>0, & 1\leq f\leq r,\\
        \lVert m_{:f}\rVert_2=0, & r<f\leq F.
    \end{cases}
\end{align}
Here, $r=0$ corresponds to the unspecialized permutation symmetric solution. \textcolor{MEM}{PSB} phases can also occur at low regularization; for simplicity, we restrict their analysis to the flipping-spike model.

\paragraph{Head-specialization phase transition in the flipping spike model.} 
We first turn to the flipping spike model defined in Sec.~\ref{model}, where a constant latent direction is learnable without head specialization while a flipping direction requires specialized heads to be used in the retrieval task. At sufficiently large regularization $\lambda \ge \lambda_{\rm tric}$, where $\lambda_{\rm tric}$ is the smallest regularization strength for which the \textcolor{MEM}{PSB} phase is absent, we find two solutions:
\begin{itemize}
    \item \textcolor{PS}{PS solution for $\alpha<\alpha_c^\mathrm{spec}(\lambda)$}: both heads implement the same attention pattern, and generalization improves with sample complexity through recovery of the constant direction ($\lVert m_{:1} \rVert_2>0$), while the flipping feature remains unrecovered ($\lVert m_{:2} \rVert_2=0$).
    \item \textcolor{SPEC}{SPEC solution for $\alpha>\alpha_c^\mathrm{spec}(\lambda)$}: the heads implement distinct attention patterns, each adapted to one sign of the flipping feature ($\lVert m_{:2} \rVert_2>0$). Recovery of this additional direction improves retrieval, with a gain in error depending on the relative signal strengths.
\end{itemize}
The multi-head attention model displays a phase transition in sample complexity from the \textcolor{PS}{PS} solution to the \textcolor{SPEC}{SPEC} solution at critical sample complexity $\alpha_c^\mathrm{spec}$. The transition is of second order, marked by non-analyticities in the overlaps and errors. 

\begin{figure}[tbp]
    \centering
    \includegraphics[width=\linewidth]{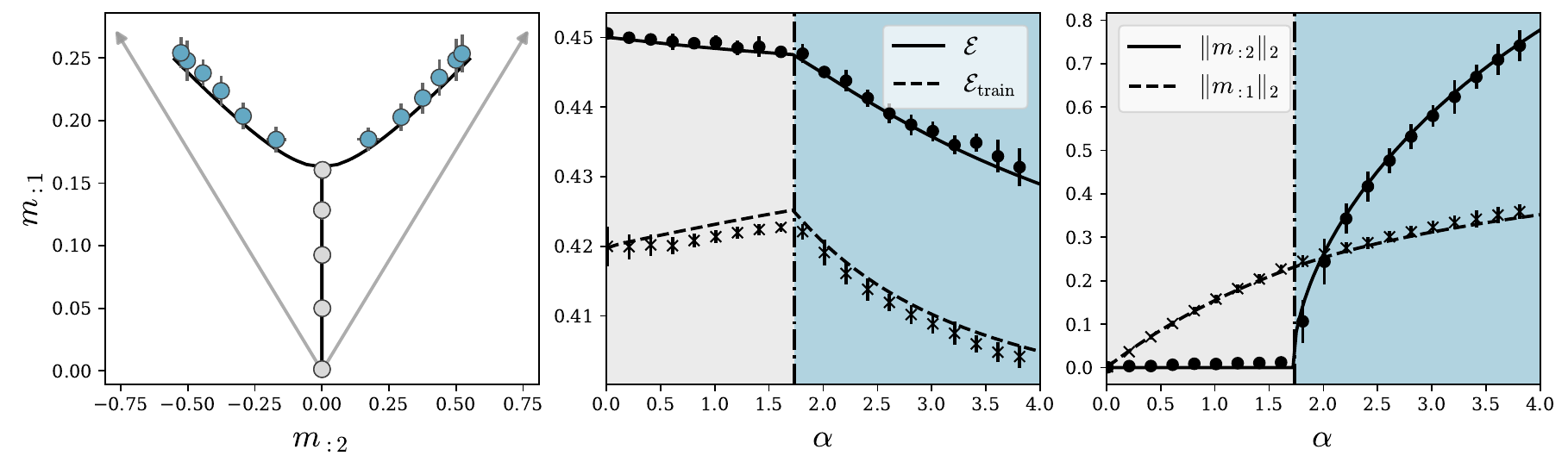}
    \caption{\textbf{Head-specialization phase transition in the flipping spike model:} $\nu_1=0.1$, $\nu_2=2.0$, $L=10$, $H=2$, $\lambda=0.2$, $D=2000$. Here and in Fig.~\ref{Fig:2} and \ref{Fig:3}, black solid lines correspond to theoretical predictions (Res.~\ref{res:self-consistent_equations}). Markers with error bars correspond to training the model (Sec.~\ref{model}) with L-BFGS from random initialization in dimension $D$ across ten seeds. \textbf{Left:} Head-feature overlaps $(m_{:2},m_{:1})$ parametrized by sample complexity. Grey arrows indicate the directions of the two possible draws of the planted spike $\hat{k}$. \textbf{Center:} Test error $\mathcal{E}$ and training error $\mathcal{E}_{\mathrm{train}}$. \textbf{Right:} Norm of head-feature overlaps $\lVert m_{:1}\rVert_2$ , $\lVert m_{:2}\rVert_2$. Grey and blue identify the permutation symmetric (\textcolor{PS}{PS}) and specialized (\textcolor{SPEC}{SPEC}) phases. Vertical lines mark the specialization threshold $\alpha_c^\mathrm{spec}$.}
    \label{Fig:1}
\end{figure}

In Fig.~\ref{Fig:1}, we compare different metrics for the lowest training loss solutions of the self-consistent equations as a function of sample complexity with the corresponding experimental results. Gradient based optimization on the ERM objective from random initializations reaches the predicted global minima, with errors and overlaps matching theoretical predictions. The transition is visible through the continuous splitting of the head-feature overlaps, the associated non-analyticity in the errors, and the onset of flipping-feature recovery at $\alpha_c^\mathrm{spec}$.

\begin{figure}[tbp]
    \centering
    \includegraphics[width=\linewidth]{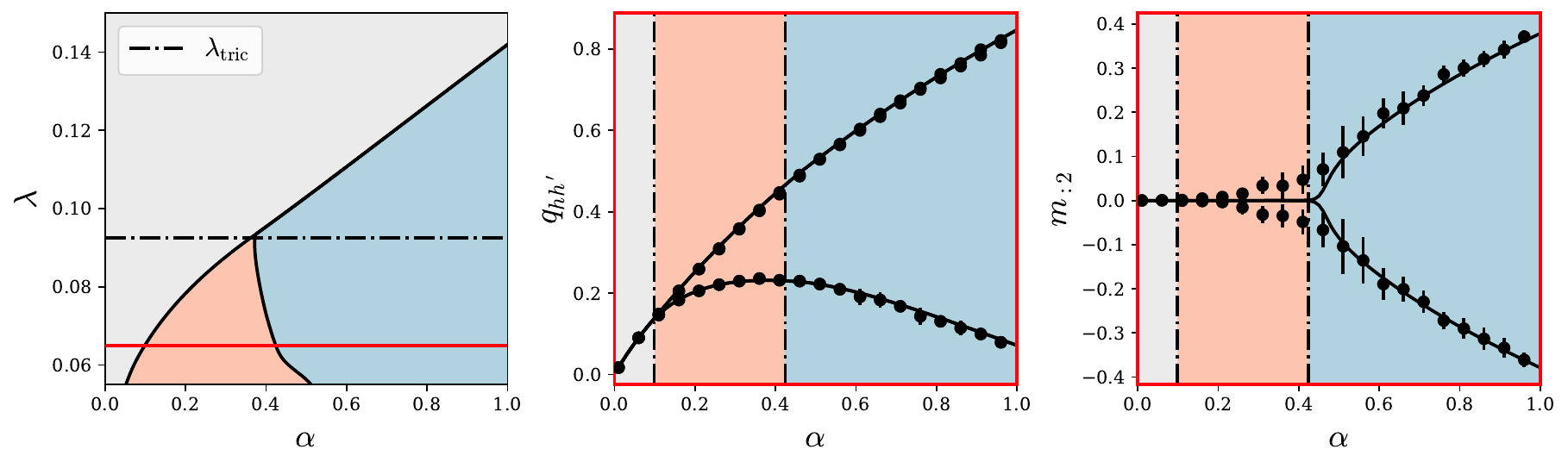}
    \caption{\textbf{Head differentiation without specialization in the flipping spike model:} $\nu_1=0.5$, $\nu_2=2.0$, $L=10$, $H=2$, $D=6000$. \textbf{Left:} Phase diagram in sample complexity $\alpha$ and regularization $\lambda$. Phase boundaries are obtained by stability analysis of the different solutions (App.~\ref{App:Susceptibility}). The red horizontal line at $\lambda=0.065$ selects the cut shown in the other panels. \textbf{Center:} Head self-overlaps $q_{11}=q_{22}$ and mutual overlaps $q_{12}=q_{21}$. \textbf{Right:} Head-feature overlaps $m_{12}$ and $m_{22}$. Grey, orange and blue identify the permutation symmetric (\textcolor{PS}{PS}), permutation symmetry broken (\textcolor{MEM}{PSB}) and specialized (\textcolor{SPEC}{SPEC}) phases, respectively. Vertical lines mark their boundaries $\alpha_c^\mathrm{psb}$ and $\alpha_c^\mathrm{spec}$.}
    \label{Fig:2}
\end{figure}

Low regularization strengths can favor fitting sample fluctuations without feature recovery. In the regime $\lambda<\lambda_{\rm tric}$, we find an additional solution to the asymptotic characterization:
\begin{itemize}
    \item \textcolor{MEM}{PSB solution for $\alpha_c^\mathrm{psb}(\lambda)<\alpha<\alpha_c^\mathrm{spec}(\lambda)$}: The two heads differ in components orthogonal to the planted directions. They have identical feature overlaps, and therefore do not recover the flipping spike.
\end{itemize}
When this solution becomes global, specialization in sample complexity is delayed, producing a re-entrant specialization phase boundary at low regularization. Figure~\ref{Fig:2} shows the resulting phase diagram and a low-regularization cut, where the \textcolor{MEM}{PSB} solution is visible in the head self-overlaps before the onset of flipping-feature recovery. A similar type of solution, noted "memorization", has been observed in the ERM of committee-machines \citep{ortiz26committee}. Here we stress that, at low regularization, the validity of the RS ansatz requires particular scrutiny. In Fig.~\ref{Fig:2} the validity of RS assumption in this regime is supported through agreement with simulations. In general, the possible need for a replica-symmetry-breaking (RSB) description needs to be investigated.

\begin{figure}[tp]
    \centering
    \includegraphics[width=\linewidth]{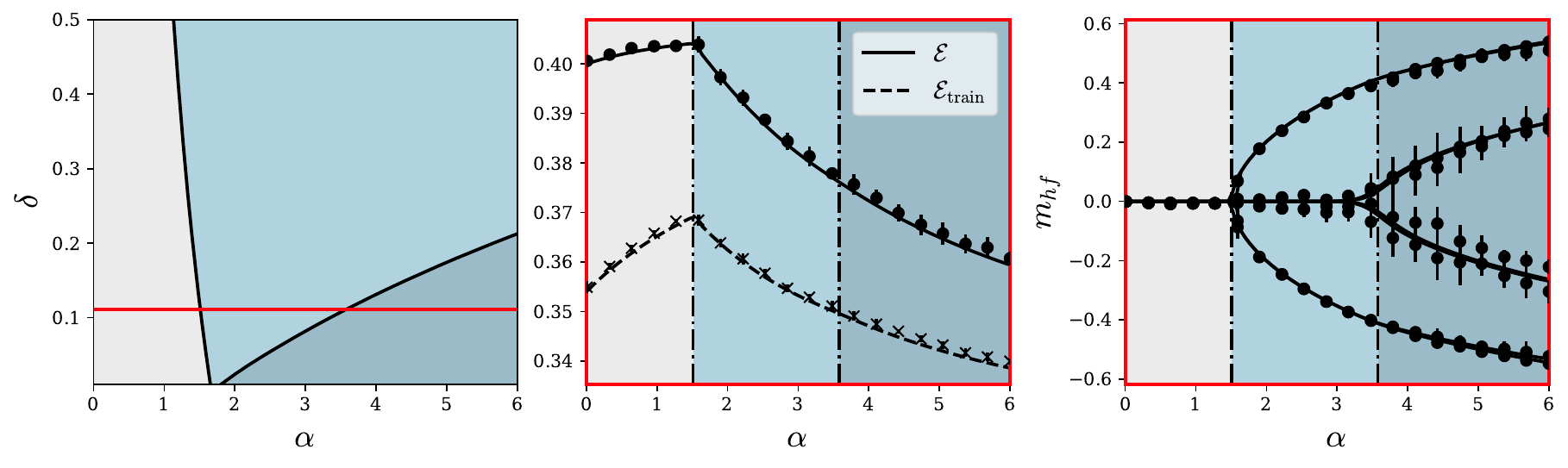}
    \caption{\textbf{Head-specialization cascade in the non-isotropic Gaussian model:} $F=2$, $\nu_\mathrm{tot}=9.0$, $L=5$, $H=4$, $\lambda=0.2$, $D=4000$. \textbf{Left:} Phase diagram in sample complexity $\alpha$ and anisotropy $\delta$. Phase boundaries are obtained by stability analysis of the different solutions (App.~\ref{App:Susceptibility}). The red horizontal line at $\delta=1/9$ corresponding to $(\nu_1, \nu_2)=(5.0, 4.0)$ selects the cut shown in the other panels. \textbf{Center:} Test error $\mathcal{E}$ and training error $\mathcal{E}_{\mathrm{train}}$. \textbf{Right:} Head-feature overlaps $m_{hf}$ for all heads and both latent features. Grey, light blue and darker blue identify the permutation symmetric (\textcolor{PS}{PS}) and successively specialized (\textcolor{SPEC}{$1$-SPEC}, \textcolor{2-SPEC}{$2$-SPEC}) phases, respectively. Vertical dash-dotted lines mark the specialization thresholds $\alpha_c^{(1)}$ and $\alpha_c^{(2)}$ along the cut.}
    \label{Fig:3}
\end{figure}

\paragraph{Head-specialization cascade and simplicity bias in the non-isotropic Gaussian model.}
We now consider the non-isotropic Gaussian model, with zero-mean feature weights of strengths $\{\nu_f\}_{f=1}^F$. For $F=2$, we parameterize the latent structure by its total signal $\nu_\mathrm{tot}$ and anisotropy $\delta$:
\begin{align}
    \nu_\mathrm{tot}=\nu_1+\nu_2,
    \qquad
    \delta=\frac{\nu_1-\nu_2}{\nu_1+\nu_2},
\end{align}
where $\delta=0$ corresponds to isotropic strengths and increasing $\delta$ favors one latent direction. We take $H=2F$, allowing one pair of heads per feature. At sufficiently large regularization we find:
\begin{itemize}
    \item \textcolor{PS}{PS solution for $\alpha<\alpha_c^{(1)}$}: no planted feature is recovered. In the absence of an easy direction, specialization is necessary to acquire the retrieval capability. The test error initially increases as the model fits uninformative sample fluctuations.
    
    \item \textcolor{SPEC}{$r$-SPEC solution for $\alpha_c^{(r)}<\alpha<\alpha_c^{(r+1)}$}: the heads recover the $r$ strongest latent features, from largest to smallest variance. Each additional specialization transition therefore recovers a weaker feature and incrementally improves retrieval.
\end{itemize}
Increasing sample complexity therefore produces a cascade of second-order specialization phase transitions at critical sample complexities  $\{\alpha_c^{(r)}\}_{r=1}^{F}$, marked by the recovery of successive latent features and associated non-analyticities in the errors. With stronger features recovered at lower sample complexity, this hierarchy constitutes a statistical form of simplicity bias in the global ERM solutions, independent of the training dynamics. In Fig.~\ref{Fig:3}, the $(\alpha,\delta)$ phase diagram illustrates this simplicity bias. At fixed anisotropy $\delta$, we compare the sample-complexity evolution of the errors and head-feature overlaps predicted by Result~\ref{res:self-consistent_equations} with numerical experiments. Gradient-based optimization from random initialization reaches the predicted global minima, with errors and overlaps matching our theoretical predictions. The Bayes-optimal baseline in App.~\ref{app:bayes_optimal} distinguishes the information-theoretic thresholds for feature recovery in the data model from the larger specialization thresholds of the attention estimator.

\section{Link between statistical transitions and dynamical emergence}\label{dynamics}
We now connect the phase transitions in sample complexity to specialization during online learning. We consider the non-isotropic Gaussian model where $(m,q)=(0,0)$ is a saddle point of the loss, and focus on the first specialization, assumed to occur only along $k_1^*$. In the right limits, we show that the sample complexity and the time for specialization are governed by the largest curvature $C\in\mathbb{R}^+$ of minus the population loss $\mathcal{E}_\mathrm{pop}$ along this feature, defined as:
\begin{align} \label{eq:hessian}
    C = \lambda_\mathrm{max}\left[-k_1^{*\top} (\nabla^2_{K,K}\mathcal E_\mathrm{pop}(0))k_1^*\right], \qquad \mathcal E_\mathrm{pop}(K) = \mathbb E_{X,y}\left[ \frac{1}{2D}\lVert y-f_K(X)\rVert_2^2 \right],
\end{align}
where $\lambda_\mathrm{max}[\cdot]$ denotes the largest eigenvalue and $\nabla^2_{K,K}$ denotes the Hessian.

To describe the dynamics, we consider the gradient-flow limit of unregularized online SGD,
\begin{align}
    \frac{\mathrm dK}{\mathrm d\tau}=-D\nabla_K\mathcal E_\mathrm{pop}(K)
\end{align}
with independent initial weights
$k_h(\tau=0)\sim\mathcal N(0,\eta^2\mathbb I_D)$. We take the limit of vanishing initialization $\eta\to0$ after $D\to\infty$, often referred to as the feature-learning limit (e.g. \cite{wu2025landscapeRelu}). \citet{sagitova2026specializationsoftmaxattentionheads} derives closed evolution equations for the order parameters and shows that specialization is governed by an exponential escape from the saddle of the head-feature overlap $m$. A self-contained derivation is provided in App.~\ref{partApp:spéGF}. We compare these dynamics with ERM (Res.~\ref{res:self-consistent_equations}) in the limit of large regularization $\lambda, \alpha\gg 1$ with $\alpha/\lambda=\Theta(1)$, in which the ERM landscape is well approximated by the population loss. A linear stability analysis (App.~\ref{partApp:spéERM}) then relates the critical sample complexity to the specialization time through the Hessian (\ref{eq:hessian}). Defining $\tau^\mathrm{spec}$ as the first time at which the specializing component reaches a fixed small amplitude $\lVert m_{:1}\rVert>\varepsilon$, for small enough 
$\varepsilon>0$ independent of $D$ and $\eta$, we obtain:
\begin{align}
    \lim_{\eta\to0^+}\lim_{D\to\infty} \frac{\tau^\mathrm{spec}}{\log(\sqrt D/\eta)} = \lim_{\lambda\to\infty} \frac{\alpha_c^\mathrm{spec}}{\lambda} = \frac{1}{C}.
\end{align}
In the considered limits the rescaled static and dynamical specialization transitions are governed by the same
instability. The rescaled time is asymptotically deterministic and independent of the initialization and of $\varepsilon$. As in online learning training, time and number of observed sequences increase together, this correspondence illustrates how a statistical phase transition can manifest itself dynamically through sudden drops in the loss at critical training times.

\section{Discussion and conclusion}\label{discussion}
The theoretical picture developed in this work suggests that abrupt and stagewise changes observed during training of multi-head attention layers can be dynamical manifestations of underlying statistical phase transitions. Solvable models make this perspective precise by identifying a limit in which specialization phase transitions are well defined, with sample complexity as a control parameter for studying the minima of the ERM objective. In our setting, this description distinguishes the statistical origins of specialization, encoded in the structure of the ERM loss landscape, from its dynamical origins. This theoretical perspective helps to clarify a possible statistical interpretation of the drops in loss that accompany abrupt changes in capabilities during the training of language models.

MLM experiments with BERT models provide strong empirical support for this statistical interpretation. Syntactic attention structure associated with head specialization emerges sharply with increasing dataset sizes, coherently with the already observed drops in the loss in training time, as suggested by the connections between ERM and online learning. While, by nature, finite-dimensional experiments do not by themselves establish the asymptotic non-analyticities that define phase transitions, these observations parallel the specialization phenomenology of the solvable model and strongly suggest a statistical phase transition interpretation of the emergence of specialized heads and SAS in BERT. For head specialization, sample complexity appears as the principled choice of control parameter to study these phase transitions. More generally, other capability changes could also have purely dynamical origins and not have a corresponding statistical characterization. 

Our results motivate further experiments to characterize specialization transitions yet more systematically. For example, the enhanced variability across seeds near the emergence of SAS observed in Fig.~\ref{Fig:0} suggests that fluctuations could help locate candidate transitions in sample complexity. This approach is motivated by the diverging sensitivity of macroscopic observables to perturbations near second-order phase transitions, a familiar phenomenon in physics. Related considerations about variability across runs are beginning to be explored in mechanistic interpretability \citep{juneja2023linear,zhao2026random}. A further test of the sample-complexity interpretation would extend our BERT experiments by varying training procedures with regularized objective and training controlled to convergence. For protocols reaching comparable ERM minima, one could examine whether SAS emerges at the same characteristic dataset size. Repeating this comparison across model sizes could help identify an effective scaling of sample complexity, while extending the analysis to individual syntactic relations could reveal distinct thresholds for the emergence of statistically more complex capabilities.

\subsection*{Reproducibility statement}
A complete derivation of all the claims is included in the appendix with references provided in the main text. All the results can be reproduced using the code provided in the repository: \url{https://github.com/SPOC-group/Attention-head-specialization}. For the experimental setup, details are provided in App.~\ref{App:Experiments}.

\subsection*{Acknowledgments}
We thank Margarita Sagitova for discussions and insights on the multi-feature spiked single-location regression model from \cite{sagitova2026specializationsoftmaxattentionheads}. We thank Pierre Marion and Naomi Saphra for insightful discussions. We acknowledge funding from the Swiss National Science Foundation grants SNSF SMArtNet (grant number 212049), the Simons Collaboration on the Physics of Learning and Neural Computation via the Simons Foundation grant ($\#$1257413 (LZ)), and the Swiss State Secretariat for Education, Research and Innovation (SERI), which funded the ERC Advanced Grant SPASM-net.

\subsection*{AI use statement}
In this work, we used generative AI tools to help formulate mathematical claims, implement numerical methods, and polish writing. We have reviewed all AI-assisted work. We used LLMs to sketch extensions of our hand-derived linear stability analysis to other solutions and to adapt the associated numerical implementations for the evaluation of critical sample complexity. We also used LLMs to help implement and reproduce a previously designed experiment using a reference paper’s repository. Finally, we used LLMs for language editing (grammar, spelling). We take responsibility for the final content of this work, including text, claims or artifacts produced with the aid of generative AI.

\bibliography{biblio_arxiv}
\bibliographystyle{abbrvnat}

\clearpage

\section*{Appendix: Table of Contents}
\appendix
\startcontents[sections]
\printcontents[sections]{l}{1}{\setcounter{tocdepth}{2}}
\newpage

\section{Derivations for the exact characterization of the ERM}
\label{App:Replica}
\subsection{Derivation of result \ref{res:self-consistent_equations} with the replica method}
\label{app:replica_derivation}

This appendix provides a derivation of result \ref{res:self-consistent_equations} based on the replica method of the statistical physics of disordered systems. The derivation is standard and closely related to \cite{ICLR2026_8c21b98a}.

\subsubsection{Setting and notation}
Recall that our goal is to characterize the minimizers of the regularized empirical risk (\ref{eq:ERM}) for the multi-head attention estimator (\ref{eq:estimator}) in the high-dimensional setting of Sec.~\ref{model}. We take $N,D\to\infty$ at fixed $\alpha=N/D$, $H$, $F$, and $L$, such that $D^{-1}K^*K^{*\top}\to\mathbb I_F$. Let $\hat K$ be the minimizer of the empirical risk:
\begin{align}\label{eq:empirical_loss}
    \mathcal L(K) &=\sum_{\mu=1}^N\ell\bigl(y_\mu,f_K(X_\mu)\bigr)+\frac{\lambda}{2}\lVert K\rVert_F^2, &\ell(y,\hat y)&=\frac{1}{2D}\lVert y-\hat y\rVert^2,
\end{align}
where $f_K$ is the estimator (\ref{eq:estimator}), and we assume $\lambda>0$. Both error and regularization contributions to the loss are extensive; we define the unregularized training error associated with $\hat K$
\begin{align}\label{eq:intensive_training_loss}
    \mathcal E_{\mathrm{train}} &=\frac{1}{N}\sum_{\mu=1}^N \ell \left(y^\mu, f_{\hat K}(X^\mu) \right).
\end{align}
 We first derive the results for generic feature weights distribution $P_\theta$, and later evaluate the remaining feature average for the flipping spike and non isotropic Gaussian models. We also consider generic activation function $\sigma:\mathbb{R}^L\rightarrow\mathbb{R}^L$, and specialize to softmax at the end of the derivation. 

\subsubsection{Gibbs formulation of the minimization problem}
To begin with, we define the Gibbs measure $d\mu_\beta$ over weights $K\in\mathbb{R}^{H\times D}$, and recall that for any observable $\phi(\hat K)$ of the minimizer $\hat K$ of (\ref{eq:ERM}),
\begin{align}\label{eq:Gibbs_measure}
    \mathbb{E}_\mathcal{D}\left[\phi(\hat K)\right] = \underset{\beta \rightarrow \infty}{\lim} \mathbb{E}_\mathcal{D}\left[\int \phi(K) d\mu_\beta \right], && d\mu_\beta = \frac{1}{\mathcal{Z}_\beta}e^{-\beta \mathcal{L}(K)} d K ,
\end{align}
where $\mathcal{Z}_\beta$ is the normalization factor also known as partition function in statistical physics:
\begin{align}
    \mathcal{Z}_\beta = \int dK \; e^{-\beta \mathcal{L}(K)}.
\end{align}
The goal of the replica method is to compute the generating function associated to the Gibbs measure (\ref{eq:Gibbs_measure}), the free entropy density
\begin{align}
    \Phi_\beta = \underset{D\rightarrow\infty}{\lim}\frac{1}{D}\;\mathbb E_\mathcal{D} \left[\log \mathcal{Z}_\beta\right],
\end{align}
building on the identity 
\begin{align}\label{eq:replica_trick}
    \mathbb E_\mathcal{D}\left[\log\mathcal Z_\beta \right] =\lim_{n\to0}\frac{\mathbb E_\mathcal{D}[\mathcal Z_\beta^n]-1}{n}.
\end{align}
Therefore, the whole problem reduces to computing the average replicated partition function:
\begin{align}\label{eq:replicated_partition_function}
     \mathbb E_\mathcal{D}\left[\mathcal Z_\beta^n\right] =\mathbb E _\mathcal{D}\left[ \int \prod_{a=1}^n dK^a \; e^{-\beta \sum_{a=1}^n \left[ \sum_{\mu=1}^N \ell (y_\mu, f_{K^a}(X_\mu)) + \frac{\lambda}{2}\lVert K^a \rVert_F^2\right]}\right].
\end{align}

\subsubsection{Replicated partition function}
With explicit dataset definitions, the partition function reads
\begin{align}
    \mathbb E_\mathcal{D}\left[\mathcal Z_\beta^n\right] =\int dP_{K^*} \prod_{a=1}^n dK^a\, e^{-\frac{\beta\lambda}{2}\sum_{a=1}^n\lVert K^a\rVert_F^2} \prod_{\mu=1}^N\mathbb E_{\epsilon_\mu,\theta_\mu,X_\mu,y_\mu\mid K^*}\left[ e^{-\beta\sum_{a=1}^n \ell\bigl(y_\mu,f_{K^a}(X_\mu)\bigr)}\right].
\end{align}
We introduce the local fields
\begin{align}\label{eq:replica_fields}
    Z^a_\mu =\frac{X_\mu K^{a\top}}{\sqrt D}\in\mathbb R^{L\times H},&& Z_\mu^* =\frac{X_\mu K^{*\top}}{\sqrt D}\in\mathbb R^{L\times F},
\end{align}
following jointly Gaussian statistics with means 
\begin{align}
    \mathbb{E}_X \left[ Z_\ell^a\right] =\delta_{\ell \epsilon}m^a\theta, && \mathbb{E}_X \left[ Z_\ell^*\right] =\delta_{\ell \epsilon}\theta,
\end{align}
and covariances 
\begin{align}
    \mathrm{Cov}\left(Z _\ell^a ,Z_{\ell '}^{b\top }\right)= \delta_{\ell \ell'}\frac{K^aK^{b\top}}{D},  && \mathrm{Cov}\left(Z _\ell^a ,Z_{\ell '}^{* }\right)= \delta_{\ell \ell'}\frac{K^aK^{*\top}}{D}, && \mathrm{Cov}\left(Z _\ell^*, Z_{\ell '}^{* }\right)= \delta_{\ell \ell'}\mathbb{I}_F.
\end{align}
Expanding the squared loss then yields the reparametrization 
\begin{align}
    \ell(y_\mu, f_{K^a}(X_\mu))\rightarrow \ell(\epsilon_\mu, Z^a_\mu) \coloneq \frac{1}{2}\sum_{\ell=1}^L \left(\delta_{\ell\epsilon_\mu} - \frac{1}{H}\sum_{h=1}^H \sigma\left(Z_{\mu, :h}^a\right)_\ell\right)^2,
\end{align}
where we used the fact that $X_\ell^\top X_{\ell'}/D\rightarrow \delta_{\ell \ell'}$ in the considered limit. Enforcing the definitions of local fields with Dirac deltas in the replicated partition function gives 
\begin{align}
    \mathbb E_\mathcal{D}\left[\mathcal Z_\beta^n\right] \propto &\int dP_{K^*}
      \prod_{a=1}^n dK^a\,
      e^{-\frac{\beta\lambda}{2}\sum_{a=1}^n\lVert K^a\rVert_F^2} \notag \\
      & \prod_{\mu=1}^N dZ_\mu^* d\hat Z_\mu^* \prod_{a=1}^n dZ_\mu^ad\hat Z_\mu^a \; \mathbb{E}_{\epsilon_\mu, \theta_\mu} \left\{e^{-\beta \sum_{a=1}^n \ell(\epsilon_\mu, Z_\mu^a)} \right.\notag \\
      &\left. \mathbb E_{X_\mu\rvert \epsilon_\mu, \theta_\mu, K^*}\left[ e^{i\sum_{\ell=1}^L \hat Z_{\mu,\ell}^{*\top}\left(Z_{\mu,\ell}^*-\frac{K^*X_{\mu, \ell}}{\sqrt{D}}\right)+i\sum_{\ell=1}^L \sum_{a=1}^n \hat Z_{\mu,\ell}^{a\top}\left(Z_{\mu,\ell}^a-\frac{K^aX_{\mu, \ell}}{\sqrt{D}}\right)}\right] \right\}\\
      \propto &\int dP_{K^*}
      \prod_{a=1}^n dK^a\,
      e^{-\frac{\beta\lambda}{2}\sum_{a=1}^n\lVert K^a\rVert_F^2} \notag \\
      & \prod_{\mu=1}^N dZ_\mu^* d\hat Z_\mu^* \prod_{a=1}^n dZ_\mu^ad\hat Z_\mu^a \; \mathbb{E}_{\epsilon_\mu, \theta_\mu} \left\{e^{-\beta \sum_{a=1}^n \ell(\epsilon_\mu, Z_\mu^a) + \theta_\mu^\top Z_{\mu, \epsilon_\mu}^* - \frac{1}{2}\theta_\mu^\top \theta_\mu}\right.\notag \\
      &\left. \mathbb E_{X_\mu \sim \mathcal{N}(0,\mathbb{I}_D)}\left[ e^{i\sum_{\ell=1}^L \hat Z_{\mu,\ell}^{*\top}\left(Z_{\mu,\ell}^*-\frac{K^*X_{\mu, \ell}}{\sqrt{D}}\right)+i\sum_{\ell=1}^L \sum_{a=1}^n \hat Z_{\mu,\ell}^{a\top}\left(Z_{\mu,\ell}^a-\frac{K^aX_{\mu, \ell}}{\sqrt{D}}\right)}\right] \right\}\\
      \propto & \int dP_{K^*}
      \prod_{a=1}^n dK^a\,
      e^{-\frac{\beta\lambda}{2}\sum_{a=1}^n\lVert K^a\rVert_F^2}\prod_{\mu=1}^N dZ_\mu^* \prod_{a=1}^n dZ_\mu^a\notag  \\& \mathbb{E}_{\epsilon_\mu, \theta_\mu} \left\{e^{-\beta \sum_{a=1}^n \ell(\epsilon_\mu, Z_\mu^a) + \theta_\mu^\top Z_{\mu, \epsilon_\mu}^* - \frac{1}{2}\theta_\mu^\top \theta_\mu} \prod_{\ell=1}^L \mathcal{N}\left( \begin{pmatrix}
          Z_{\mu,\ell}^*\\Z_{\mu,\ell}
      \end{pmatrix}; 0, \begin{pmatrix}
          \mathbb I_F & m^\top \\ m & Q
      \end{pmatrix}\right) \right\}
\end{align}
where we introduced the summary statistics 
\begin{align}
    m^a =\frac{K^a K^{*\top}}{D}\in\mathbb{R}^{H\times F }, && Q^{ab}=\frac{K^a K^{b \top}}{D}\in\mathbb{R}^{H\times H }.
\end{align}
with $Z_{\mu,\ell}=\{Z_{\mu,\ell}^a\}_{a=1}^n$, $m=\{m^a\}_{a=1}^n$, $Q=\{Q^{ab}\}_{a,b=1}^n$. Enforcing the definitions of $m^a$ and $Q^{ab}$ with Dirac functions brings the replicated partition function in the following form:
\begin{align}\label{eq:decomposed_replicated_function}
    \mathbb{E}_\mathcal{D} \left[\mathcal{Z}_\beta^n \right] \propto & \int \prod_{a=1}^ndm^a d\hat m^a  \prod_{a,b=1}^ndQ^{ab} d\hat Q^{ab} \; \underbrace{e^{-D\sum_{a=1}^n\mathrm{Tr}(\hat m^{a\top}m^a) + \frac{D}{2}\sum_{a,b=1}^n \mathrm{Tr}(\hat Q^{ab\top}Q^{ab})}}_{e^{nD\Psi_t}}\notag\\
    & \underbrace{\int  dP_{K^*} \prod_{a=1}^ndK^a e^{- \frac{\beta \lambda}{2}\sum_{a=1}^n\lVert K^a\rVert_F^2 + \sum_{a=1}^n \mathrm{Tr}(\hat m^{a\top}K^aK^{*\top}) - \frac{1}{2}\sum_{a,b=1}^n \mathrm{Tr}(\hat Q^{ab}K^bK^{a\top})}}_{e^{nD\Psi_K}}\notag \\
    & \underbrace{ \mathbb{E}_{\epsilon, \theta}\left\{ \mathbb{E}_{Z^*,\{Z^a\}_{a=1}^n}\left[  e^{-\beta \sum_{a=1}^n \ell(\epsilon, Z^a) + \theta^\top Z_\epsilon^* - \frac{1}{2}\theta^\top \theta }\right]\right\}^{\alpha D}}_{e^{n\alpha D \Psi_y}}.
\end{align}
We decomposed the replicated free entropy into the trace $\Psi_t$, entropic $\Psi_K$ and energetic $\Psi_y$ potentials. In the high-dimensional limit where $D\to\infty$ while $\alpha =N/D$ stay finite, the integral (\ref{eq:decomposed_replicated_function}), written explicitly as 
\begin{align}
    \mathbb{E}_\mathcal{D} \left[\mathcal{Z}_\beta^n \right] \propto  \int \prod_{a=1}^ndm^a d\hat m^a  \prod_{a,b=1}^ndQ^{ab} d\hat Q^{ab} e^{D\Phi_\beta^{(n)}\left(m^a, \hat m^a, Q^{ab}, \hat Q^{ab}\right)},
\end{align}
concentrate around the values of the overlaps that extremize the finite temperature free entropy $\Phi_\beta^{(n)}(m^a,\hat m^a, Q^{ab}, \hat Q^{ab})=n\left(\Psi_t+\Psi_K+\alpha\Psi_y\right) $, and can be evaluated using a Laplace saddle-point approximation. 

\subsubsection{Replica-symmetric ansatz}
We have thus rephrased the analysis of the measure (\ref{eq:Gibbs_measure}) as a optimization problem over the order parameters $\{m^a, Q^{ab}\}$, and the associated conjugate variables. However, these still represent $2nHF + 2n^2H^2$ variables with $n\rightarrow0$. In order to make progress, we assume that the saddle point is of replica-symmetric (RS) form
\begin{align}
    &m^a = m \in\mathbb{R}^{H\times F }, && \hat m^a = \hat m \in\mathbb{R}^{H\times F }, \\
    &Q^{ab} = q\delta_{a\neq b} + Q \delta_{ab} \in \mathbb{R}^{H\times H }, && \hat Q^{ab} = -\hat q\delta_{a\neq b} + \hat Q \delta_{ab} \in \mathbb{R}^{H\times H } . 
\end{align}
The RS ansatz holds in a number of machine learning settings, notably for convex problems and Bayes-optimal settings. In the present setting, since the empirical loss (\ref{eq:empirical_loss})
is non-convex, the RS ansatz constitutes a heuristic technical assumption of our analysis. 

Under this ansatz, we can factorize the replicas by introducing random Gaussian variables $\Xi= \{\xi_\ell\}_{\ell=1}^L$ with $\xi_\ell \sim \mathcal{N}(0, \mathbb{I}_H)$. With quadratic regularization, we can perform the integration over the weights $K$ in the entropic term. Introducing the variance order parameter 
\begin{align}
    V = Q-q, && \hat V = \hat Q + \hat q,
\end{align}
and taking the limit $n\rightarrow0$, the free entropy can then be expressed as 
\begin{align}\label{eq:finite_temperature_replica_free_entropy}
    &\Phi_\beta = \underset{n\rightarrow0}{\lim} \frac{1}{n}\underset{m,\hat m, q, \hat q, V, \hat V}{\mathrm{extr}}\left\{\Phi^{(n)}_\beta(m^a,\hat m^a, Q^{ab}, \hat Q^{ab})\right\}=\underset{m,\hat m, q, \hat q, V, \hat V}{\mathrm{extr}} \left\{ \Psi_t + \Psi_K + \alpha \Psi_y \right\}  
\end{align}
with 
\begin{align}
    &\Psi_t = -\mathrm{Tr}(\hat m^\top m ) + \frac{1}{2} \mathrm{Tr}\left[\hat V (q+V)- \hat q V \right]\\
    & \Psi_K = -\frac{1}{2}\log \left[\det(\beta \lambda \mathbb{I}_H+ \hat V ) \right] + \frac{1}{2}\mathrm{Tr}\left[ (\beta \lambda \mathbb{I}_H + \hat V )^{-1}( \hat q + \hat m \hat m^\top )\right]\\
    & \Psi_y = \mathbb{E}_{\Xi, \epsilon, \theta} \left\{ \log \left[ \int dZ e^{-\beta \ell(\epsilon, Z)}\frac{e^{- \frac{1}{2}\sum_{\ell=1}^L (Z_\ell - \omega_\ell)^\top V^{-1}(Z_\ell - \omega_\ell)}}{\det(2\pi V)^{L/2}}\right]\right\}
\end{align}
where we introduced the local fields $\Omega=\{\omega_\ell\}_{\ell=1}^L$ with $\omega_\ell \coloneq \delta_{\ell\epsilon}m\theta+ q^{1/2}\xi_\ell$ and we omit additive constants in the potentials. Note that these expressions do not depend on $K^*$ anymore; the planted directions enter only through $K^*K^{*\top}/D\to\mathbb I_F$, such that no further specification or averaging of their distribution is required.

\subsubsection{Zero temperature limit}
We now take the limit $\beta \to \infty$, rescaling 
\begin{align}
    \hat m \to \beta \hat m , && \hat q \to \beta^2 \hat q , && \hat V \to \beta \hat V, && V \to \beta^{-1}V.
\end{align}
The potentials and free entropy reduce to 
\begin{align}\label{eq:generic_replica_free_entropy}
    \Phi = \lim_{\beta\to \infty } \frac{1}{\beta}\Phi_\beta = \underset{m,\hat m, q, \hat q, V, \hat V}{\mathrm{extr}} \left\{ \Psi_t + \Psi_K + \alpha \Psi_y \right\}
\end{align}
with 
\begin{align}
    &\Psi_t = - \mathrm{Tr}(\hat m^\top m) + \frac{1}{2}\mathrm{Tr}(\hat Vq - \hat q V) \\
    &\Psi_K = \frac{1}{2}\mathrm{Tr}\left[(\lambda \mathbb I_H + \hat V)^{-1}(\hat q + \hat m \hat m^\top) \right]\\
    & \Psi_y = -\mathbb{E}_{\Xi, \epsilon, \theta} \left[ \mathcal{M}(\epsilon, \theta, \Xi)\right]
\end{align}
where the Moreau envelope is defined as 
\begin{align}\label{eq:generic_replica_free_entropy_moreau}
    \mathcal{M}(\epsilon, \theta, \Xi)=\inf_{Z\in\mathbb{R}^{L\times H}}\left\{ \ell(\epsilon, Z) + \frac{1}{2}\sum_{\ell=1}^L (Z_\ell-\omega_\ell)^\top V^{-1}(Z_\ell-\omega_\ell)\right\}.
\end{align}
\subsubsection{Replica free entropy}
Up to this point, the derivation can be carried generally with respect to the activation function and low dimensional distributions of the data model. One already reached a low dimensional characterization and can in principle state the final result generally. For clarity, we state the results for the particular setting studied. Specifying the distribution of $\epsilon$ and using permutation equivariance of softmax with respect to the tokens to fix $\epsilon=1$, we can simplify further the equations (\ref{eq:generic_replica_free_entropy}-\ref{eq:generic_replica_free_entropy_moreau}). One finally reaches an expression for the replica free entropy:
{\small
\begin{align}\label{eq:replica_free_entropy}
    \Phi = \underset{m,\hat m, q, \hat q, V, \hat V}{\mathrm{extr}} \left\{\frac{1}{2}\mathrm{Tr}(\hat Vq - \hat q V)- \mathrm{Tr}(\hat m^\top m)+ \frac{1}{2}\mathrm{Tr}\left[(\lambda \mathbb I_H + \hat V)^{-1}(\hat q + \hat m \hat m^\top) \right]- \alpha\mathbb{E}_{\Xi, \theta} \left[ \mathcal{M}(\theta, \Xi)\right]\right\}
\end{align}}where we defined the local fields $\omega_\ell = \delta_{\ell1}m\theta + q^{1/2}\xi_\ell$ with $\xi_\ell\sim\mathcal{N}(0,\mathbb{I}_H)$ and the Moreau envelope
{\small
\begin{align}\label{eq:moreau}
    \mathcal{M}(\theta, \Xi)=\inf_{Z\in\mathbb{R}^{L\times H}}\left\{ \frac{1}{2}\sum_{\ell=1}^L\left(\delta_{\ell1} - \frac{1}{H}\sum_{h=1}^H \mathrm{softmax}(Z_{:h})_\ell\right)^2 + \frac{1}{2}\sum_{\ell=1}^L \left(Z_\ell-\omega_\ell\right)^\top V^{-1}\left(Z_\ell-\omega_\ell\right)\right\}.
\end{align}}

\subsubsection{Saddle-point equations}
The extremization for the free entropy (\ref{eq:replica_free_entropy}) obtained by setting the gradient of its potential to zero yields the following system of self-consistent equations on the summary statistics:
\begin{align}\label{eq:saddle_point_equations_app}&
    \begin{cases}
        m=(\lambda \mathbb{I}_H+\hat V)^{-1}\hat m\\ 
        q=(\lambda \mathbb{I}_H+\hat V)^{-1}(\hat m \hat m^\top+\hat q)(\lambda\mathbb{I}_H+\hat V)^{-1}\\
        V = (\lambda\mathbb{I}_H+\hat V)^{-1}
    \end{cases}\\&
    \begin{cases}
        \hat m=\alpha \mathbb{E}_{\theta, \Xi}\left[V^{-1}\left( \mathrm{prox}(\theta, \Xi)_1 - \omega_1\right)\theta^\top\right]\\
        \hat q=\alpha \mathbb{E}_{\theta, \Xi}\left[V^{-1}\left(\sum_{\ell=1}^L\left( \mathrm{prox}(\theta, \Xi)_\ell - \omega_\ell\right)^{\otimes 2} \right)V^{-1}\right]\\
        \hat V=-\alpha \mathbb{E}_{\theta, \Xi}\left[V^{-1}\left(\sum_{\ell=1}^L\left(\mathrm{prox}(\theta, \Xi)_\ell - \omega_\ell \right)\xi_\ell^\top \right)q^{-1/2} \right]
    \end{cases}
\end{align}
where the proximals $\{\mathrm{prox}(\theta,\Xi)_\ell\}_{\ell=1}^L$ refer to the minimizers of the Moreau envelope $\mathcal{M}(\theta, \Xi)$ (\ref{eq:moreau}).

Note that the update equation for $\hat V$ contains a term $q^{-1/2}$ that could become singular, in particular on the PS branch. This apparent singularity is removable using integration by parts
\begin{align}
    \hat V =-\alpha\,\mathbb E_{\theta,\Xi} \left[ \sum_{\ell=1}^L V^{-1} \left( \frac{\partial\,\mathrm{prox}(\theta,\Xi)_\ell} {\partial\omega_\ell} -\mathbb I_H \right)\right],
\end{align}
and we use this formulation to improve the stability of the numerical implementation of state evolution for the non-isotropic Gaussian model.

\subsubsection{Test error}
The generalization performance is measured by the test error 
\begin{align}\label{eq:original_test_error}
    \mathcal{E} = \lim_{D\to \infty }\mathbb{E}_\mathcal{D}\left\{\mathbb{E}_{X,y\rvert K^*}\left[ \frac{1}{2D}\lVert y - f_{\hat K}\left( X\right)\rVert^2\right]\right\},
\end{align}
where the test samples $\{X,y\}$ are independent of the training set $\mathcal{D}$, conditionally on the planted spikes $K^*$. Introducing the local fields $X\hat K^\top/\sqrt{D}$ and expliciting the expression for the data model, the test error admits an asymptotic reparameterization in terms of the summary statistics characterized by (\ref{eq:saddle_point_equations_app}):
\begin{align}
    \mathcal{E} =\frac{1}{2}\mathbb{E}_{\theta,\Xi}\left[\sum_{\ell=1}^L\left( \delta_{\ell 1}- \frac{1}{H}\sum_{h=1}^H\mathrm{softmax}(\Omega_{:h})_\ell\right)^2 \right]
\end{align}
where, as before, $\theta\sim P_\theta$, $\Xi=\{\xi_\ell\}_{\ell=1}^L$ with $\xi_\ell\sim\mathcal{N}(0,\mathbb I_H)$, $\Omega = \{\omega_\ell\}_{\ell=1}^L$ $\omega_\ell = \delta_{\ell 1}m\theta + q^{1/2}\xi_{\ell}$, and we used token permutation equivariance of the softmax activation. 

\subsubsection{Training error}
As noted in \citet{NEURIPS2024_3fefebc2}, the unregularized training error (\ref{eq:intensive_training_loss}) can be obtained through the free energy $-\Phi$ at zero temperature:
\begin{equation}
    \mathcal{E}_\mathrm{train} = -\frac{1}{\alpha}\Phi -\frac{\lambda}{2\alpha}\mathrm{Tr}(q)= \mathbb{E}_{\theta, \Xi} \left[\mathcal{M}(\theta, \Xi)\right] - \frac{1}{2\alpha}\mathrm{Tr}(\hat q V),
\end{equation}
where again, $\Xi=\{\xi_\ell\}_{\ell=1}^L$ with $\xi_\ell\sim\mathcal{N}(0,\mathbb I_H)$, and $\mathcal{M}(\theta, \Xi)$ denotes the Moreau envelope (\ref{eq:moreau}). This can also be obtained directly from the relation $\mathcal{L}_\mathrm{train}=-\lim_{\beta \to \infty }\partial_\beta \log \mathcal{Z}_\beta $, where the derivative is taken on the finite temperature free entropy (\ref{eq:finite_temperature_replica_free_entropy}) before the limit. 

\subsection{Model-specific saddle-point equations}\label{app:model_specific_saddle-point_equations}
We now evaluate the remaining feature averages for the two choices of $P_\theta$ defining the models of Sec.~\ref{model}.
\subsubsection{Flipping spike model}
In the case of the flipping spike distribution where $\theta_1 = \sqrt{\nu_1}$ and $\theta_2\sim \mathrm{Unif}(\{-\sqrt{\nu_2}, \sqrt{\nu_2}\})$ we can define 
\begin{align}
    P_\theta = \frac{1}{2}(\delta_{\theta_+} + \delta_{\theta_-}), && \theta_\pm = \begin{pmatrix}
        \sqrt{\nu_1} \\ \pm \sqrt{\nu_2}
    \end{pmatrix}.
\end{align}
The random variable is discrete and averaging over $\theta$ leads to the saddle-point equations:
\begin{align}&
    \begin{cases}
        m=(\lambda \mathbb{I}_H+\hat V)^{-1}\hat m\\ 
        q=(\lambda \mathbb{I}_H+\hat V)^{-1}(\hat m \hat m^\top+\hat q)(\lambda\mathbb{I}_H+\hat V)^{-1}\\
        V = (\lambda\mathbb{I}_H+\hat V)^{-1}
    \end{cases}\\&
    \begin{cases}
        \hat m=\frac{\alpha}{2} \sum_{s=\pm 1}\mathbb{E}_{\Xi}\left[V^{-1}\left( \mathrm{prox}(\Xi)_1^{(s)} - \omega_1^{(s)}\right)\theta_s^\top\right]\\
        \hat q=\frac{\alpha}{2} \sum_{s=\pm 1} \mathbb{E}_{ \Xi}\left[V^{-1}\left(\sum_{\ell=1}^L\left( \mathrm{prox}(\Xi)_\ell^{(s)} - \omega_\ell^{(s)}\right)^{\otimes 2} \right)V^{-1}\right]\\
        \hat V=-\frac{\alpha}{2} \sum_{s=\pm 1} \mathbb{E}_{ \Xi}\left[V^{-1}\left(\sum_{\ell=1}^L\left(\mathrm{prox}(\Xi)_\ell^{(s)} - \omega_\ell^{(s)} \right)\xi_\ell^\top \right)q^{-1/2} \right]
    \end{cases}
\end{align}
where we define the local fields $\Omega^{(s)}=\{\omega_\ell^{(s)}\}_{\ell=1}^L$ as $\omega_\ell^{(s)}\coloneq \delta_{\ell1}m\theta_s + q^{1/2}\xi_\ell$ and with $\xi\sim \mathcal{N}(0,\mathbb I_H)$ the  proximals $\{\mathrm{prox}(\Xi)_\ell^{(s)}\}_{\ell=1}^L$ refer to the minimizers of the Moreau envelope
{\small
\begin{align}
    \mathcal{M}(\Xi)^{(s)}=\inf_{Z\in\mathbb{R}^{L\times H}}\left\{ \frac{1}{2}\sum_{\ell=1}^L\left(\delta_{\ell1} - \frac{1}{H}\sum_{h=1}^H \mathrm{softmax}(Z_{:h})_\ell\right)^2 + \frac{1}{2}\sum_{\ell=1}^L \left(Z_\ell-\omega_\ell^{(s)}\right)^\top V^{-1}\left(Z_\ell-\omega_\ell^{(s)}\right)\right\}.
\end{align}}
The test error then becomes 
\begin{align}
    \mathcal{E} =\frac{1}{4}\sum_{s=\pm 1}\mathbb{E}_{\Xi}\left[\sum_{\ell=1}^L\left( \delta_{\ell 1}- \frac{1}{H}\sum_{h=1}^H\mathrm{softmax}(\Omega_{:h}^{(s)})_\ell\right)^2 \right],
\end{align}
and the train error
\begin{equation}
    \mathcal{E}_\mathrm{train} = \frac{1}{2}\sum_{s=\pm 1} \mathbb{E}_{\Xi} \left[\mathcal{M}(\Xi)^{(s)}\right] - \frac{1}{2\alpha}\mathrm{Tr}(\hat q V) .
\end{equation}

\subsubsection{Non-isotropic Gaussian model}
In the case of the non-isotropic Gaussian distribution we have
\begin{align}
    \theta\sim\mathcal{N}(0,\Lambda),
    && \Lambda=\mathrm{diag}(\nu_1,\ldots,\nu_F).
\end{align}
Dependencies on $\theta$ enter the proximals only through the local fields $\Omega = \{\omega_\ell\}_{\ell=1}^L$ and we have
\begin{align}
    \mathbb{E}[\theta\mid\Omega]
    &=\Lambda m^\top(q+m\Lambda m^\top)^{-1}\omega_1,
    &
    \mathbb{E}[\xi_\ell\mid\Omega]
    &=q^{1/2}(q+\delta_{\ell1}m\Lambda m^\top)^{-1}\omega_\ell.
\end{align}
Averaging over $(\theta,\Xi)$ conditionally on the local fields then leads to the saddle-point equations:
\begin{align}&
    \begin{cases}
        m=(\lambda\mathbb{I}_H+\hat V)^{-1}\hat m\\
        q=(\lambda\mathbb{I}_H+\hat V)^{-1}
        (\hat m\hat m^\top+\hat q)
        (\lambda\mathbb{I}_H+\hat V)^{-1}\\
        V=(\lambda\mathbb{I}_H+\hat V)^{-1}
    \end{cases}\\&
    \begin{cases}
        \hat m=\alpha\mathbb{E}_{\Omega}\left[
        V^{-1}\left(\mathrm{prox}(\Omega)_1-\omega_1\right)
        \omega_1^\top\right](q+m\Lambda m^\top)^{-1}m\Lambda\\
        \hat q=\alpha\mathbb{E}_{\Omega}\left[
        V^{-1}\left(\sum_{\ell=1}^L
        \left(\mathrm{prox}(\Omega)_\ell-\omega_\ell\right)^{\otimes2}
        \right)V^{-1}\right]\\
        \hat V=-\alpha\mathbb{E}_{\Omega}\left[
        V^{-1}\left(\sum_{\ell=1}^L
        \left(\mathrm{prox}(\Omega)_\ell-\omega_\ell\right)
        \omega_\ell^\top(q+\delta_{\ell1}m\Lambda m^\top)^{-1}\right)\right]
    \end{cases}
\end{align}
where $\Omega = \{\omega_\ell\}_{\ell=1}^L$ with $\omega_\ell\sim \mathcal{N} \left(0,q+\delta_{\ell1}m\Lambda m^\top\right)$ and $\{\mathrm{prox}(\Omega)_\ell\}_{\ell=1}^L$
refer to the minimizers of the Moreau envelope
{\small
\begin{align}
    \mathcal{M}(\Omega)
    =\inf_{Z\in\mathbb{R}^{L\times H}}\left\{
    \frac{1}{2}\sum_{\ell=1}^L
    \left(\delta_{\ell1}
    -\frac{1}{H}\sum_{h=1}^H
    \mathrm{softmax}(Z_{:h})_\ell\right)^2
    +\frac{1}{2}\sum_{\ell=1}^L
    (Z_\ell-\omega_\ell)^\top
    V^{-1}(Z_\ell-\omega_\ell)
    \right\}.
\end{align}}
The test error then becomes
\begin{align}
    \mathcal{E} =\frac{1}{2}\mathbb{E}_{\Omega}\left[ \sum_{\ell=1}^L\left(\delta_{\ell1} -\frac{1}{H}\sum_{h=1}^H \mathrm{softmax}(\Omega_{:h})_\ell \right)^2\right],
\end{align}
and the train error
\begin{equation}
    \mathcal{E}_{\mathrm{train}} =\mathbb{E}_{\Omega}\left[\mathcal{M}(\Omega)\right] -\frac{1}{2\alpha}\mathrm{Tr}(\hat qV).
\end{equation}

\subsection{Bayes-optimal}
\label{app:bayes_optimal}
In this section, we characterize the Bayes-optimal (BO) estimator \(\hat y_{\mathrm{BO}}(X,\mathcal D)\), which minimizes the expected test squared error among all predictors based on the training dataset \(\mathcal D\) and the test input \(X\). The data-generating model and priors are assumed known, while the planted spikes must be inferred from \(\mathcal D\). The Bayes risk \(\mathcal E_{\mathrm{BO}}\) therefore acts as a benchmark providing a lower bound on the expected test error achievable by the attention model. Inferring the planted spikes can be reduced to low-rank matrix estimation, studied in \cite{lesieur2017constrained}. For completeness, we provide a self-contained characterization of the Bayes-optimal estimator in our setting. Given the training dataset $\mathcal D=\{X_\mu,y_\mu\}_{\mu=1}^N$, the expected test square error of the BO is 
\begin{align}\label{Eq:BO_error}
    \mathcal{E}_\mathrm{BO} = \mathbb{E}_{X, y, \mathcal{D}}\left[\frac{1}{2D}\lVert y - \hat y_\mathrm{BO}(X, \mathcal{D})\rVert^2_2 \right],
\end{align}
where the estimator $\hat y_\mathrm{BO}(X, \mathcal{D})$ is the posterior mean under the probabilistic model of Sec.~\ref{model}:
\begin{align}
    \hat y_\mathrm{BO}(X, \mathcal{D}) = \mathbb{E}_y\left[y\mid X,\mathcal{D}\right] = \mathbb{E}_\epsilon\left[X_\epsilon\mid X,\mathcal{D}\right] =\sum_{\ell=1}^L P(\ell = \epsilon \mid  X, \mathcal{D})X_\ell.
\end{align}
We can compute it using the Bayes formula
\begin{align}
    P(\ell = \epsilon \mid X, \mathcal{D}) = \frac{P(\ell = \epsilon, X \mid \mathcal{D})}{P( X \mid \mathcal{D})} = \frac{P(\ell = \epsilon, X \mid \mathcal{D})}{\sum_{\ell' = 1}^L P(\ell'=\epsilon,  X \mid \mathcal{D})},
\end{align}
where the numerator is 
\begin{align}
    P(\ell = \epsilon, X \mid \mathcal{D}) &= \int dK d\theta \; P(X, \ell=\epsilon, \theta, K \mid \mathcal{D}) \notag \\
    &= \int dK d\theta \; P(X \mid  \ell=\epsilon, \theta, K) P(\ell =\epsilon)P(\theta ) P(K\mid \mathcal{D})
\end{align}
and, given the data generation process,
\begin{align}
    P(X\mid \ell = \epsilon, \theta, K) = \frac{1}{(2\pi)^{LD/2}}e^{X_\ell^\top \frac{K^\top \theta}{\sqrt{D}}-\frac{1}{2}\lVert\frac{ K^\top \theta}{\sqrt{D}}\rVert_2^2}\prod_{\ell'=1}^L e^{-\frac{1}{2}\lVert X_{\ell'}\rVert_2^2}.
\end{align}
Putting everything together, we get 
\begin{align}\label{Eq:BO_estimator}
    \hat y_\mathrm{BO}(X, \mathcal{D}) = \sum_{\ell=1}^L \frac{\mathbb{E}_{K\mid \mathcal{D}}\mathbb{E}_\theta \left[\mathrm{exp}\left( X_\ell^\top (K^\top \theta/\sqrt{D}) -  \lVert K^\top \theta/\sqrt{D} \rVert_2^2/2\right)\right]}{\sum_{\ell'=1}^L \mathbb{E}_{K\mid \mathcal{D}}\mathbb{E}_\theta \left[\mathrm{exp}\left( X_{\ell'}^\top (K^\top \theta/\sqrt{D}) -  \lVert K^\top \theta/\sqrt{D} \rVert_2^2/2\right)\right]}X_\ell.
\end{align}
The remaining steps to obtain an expression for the estimator concern the average over the  posterior $P(K\mid \mathcal{D})$. We have
\begin{align}
    P(K\mid \mathcal{D}) &= \frac{P_K(K)P(\mathcal{D}\mid K)}{\int dK' P_K(K')P(\mathcal{D}\mid K')},
\end{align}
and using marginalization to isolate the data generating process,
\begin{align}
    P(\mathcal{D}\mid K)&=\prod_{\mu = 1}^N \sum_{\ell=1}^L\int d\theta_\mu P_\theta(\theta_\mu)P(\ell=\epsilon_\mu)P(X_\mu, y_\mu \mid K, \theta_\mu, \epsilon_\mu) \notag \\
    & = \frac{\mathrm{exp}\left(-\frac{1}{2}\sum_{\mu, \ell}\lVert X_{\mu,\ell}\rVert_2^2 \right)}{L^N (2\pi)^{NLD/2}}\prod_{\mu = 1}^N \mathbb{E}_{\theta_\mu}\!\left[\mathrm{exp}\!\left(\frac{\theta_\mu^\top K}{\sqrt{D}}y_\mu - \frac{\theta_\mu^\top KK^\top \theta_\mu}{2D} \right)\right].
\end{align}
Terms independent of $K$ simplify in the numerator and denominator of the posterior, giving
\begin{align}
    P(K\mid \mathcal{D}) = \frac{P_K(K)}{\mathcal{Z}(\mathcal{D})} \prod_{\mu = 1}^N \mathbb{E}_{\theta_\mu}\!\left[\mathrm{exp}\!\left(\frac{\theta_\mu^\top K}{\sqrt{D}}y_\mu - \frac{\theta_\mu^\top KK^\top \theta_\mu}{2D}\right) \right],
\end{align}
where we introduced the reduced partition function
\begin{align}
    \mathcal{Z}(\mathcal{D}) = \int d K\;  P_K(K) \prod_{\mu = 1}^N \mathbb{E}_{\theta_\mu}\!\left[\mathrm{exp}\!\left(\frac{\theta_\mu^\top K}{\sqrt{D}}y_\mu - \frac{\theta_\mu^\top KK^\top \theta_\mu}{2D} \right)\right].
\end{align}
To compute the summary statistics of $P(K\mid \mathcal{D})$ we again resort to the replica method, computing the free entropy. The replicated partition function reads
\begin{align}
    \mathbb{E}_\mathcal{D}\left[\mathcal{Z}^n\right] =& \int dK^* P_K(K^*)\prod_{a=1}^n  dK^a P_K(K^a) \notag \\ &\prod_{\mu = 1}^N \mathbb{E}_{\theta_\mu^*, y_\mu\mid K^*}\!\left[\prod_{a=1}^n \mathbb{E}_{\theta_\mu^a}\!\left[ \mathrm{exp}\!\left(\frac{\theta_\mu^{a\top} K^a}{\sqrt{D}}y_\mu - \frac{\theta_\mu^{a\top} K^aK^{a\top} \theta_\mu^a}{2D} \right)\right] \right] \notag \\
    =& \int dK^* P_K(K^*)\prod_{a=1}^n  dK^a P_K(K^a) \notag \\ & \prod_{\mu = 1}^N \mathbb{E}_{\theta_\mu^*,\{\theta_\mu^a\}_{a=1}^n}\!\left[ \mathrm{exp}\!\left(\sum_a\frac{\theta_\mu^{a\top}K^aK^{*\top}\theta^*_\mu}{D} + \sum_{a<b}\frac{\theta_\mu^{a\top}K^aK^{b\top}\theta^b_\mu}{D} \right)\right] 
\end{align}
where in the second line we used the fact that $y_\mu = K^{*\top}\theta_\mu^*/\sqrt{D} + z_\mu$ with $z_\mu \sim \mathcal{N}(0,\mathbb{I}_D)$ and averaged over the gaussian noise. We introduce the summary statistics 
\begin{align}
    m^a = \frac{K^aK^{*\top }}{D}\in \mathbb{R}^{F\times F}, &&   Q^{ab} = \frac{K^aK^{b\top }}{D}\in \mathbb{R}^{F\times F}.
\end{align}
Assuming distinct signal strengths in the Gaussian model, these overlaps are defined with a consistent choice of signs for the latent spikes. Enforcing their definitions with Dirac deltas brings the replicated partition function in the following form:
\begin{align}
     \mathbb{E}_\mathcal{D}\left[\mathcal{Z}^n\right] = & \int \prod_{a=1}^n dm^a d\hat m^a \prod_{a,b=1}^ndQ^{ab}d\hat Q^{ab} \; \underbrace{e^{-D\sum_{a=1}^n \mathrm{Tr}(\hat m^{a\top}m^a)+\frac{D}{2}\sum_{a,b=1}^n\mathrm{Tr}(\hat Q^{ab}Q^{ab})}}_{e^{nD\Psi_t}} \notag \\
     & \underbrace{\int dK^* P_K(K^*)\prod_{a=1}^n dK^a P_K(K^a)e^{\sum_{a=1}^n \mathrm{Tr}(\hat m^{a\top}K^aK^{*\top})- \frac{1}{2}\sum_{a,b=1}^n\mathrm{Tr}(\hat Q^{ab}K^aK^{b\top})}}_{e^{nD\Psi_K}} \notag \\
     & \underbrace{\left[ \int d\theta^* P_\theta(\theta^*)\prod_{a=1}^n d\theta^a P_\theta(\theta^a) e^{\sum_{a=1}^n \theta^{a\top}m^a\theta^* + \sum_{a<b}\theta^{a\top}Q^{ab}\theta^{b} }\right]^{\alpha D}}_{e^{n\alpha D\Psi_y}}
\end{align}
We decomposed the replicated free entropy into the trace $\Psi_t$, entropic $\Psi_K$ and energetic $\Psi_y$ potentials. One again assumes the replica symmetric ansatz 
\begin{align}
    &m^a = m , && \hat m^a = \hat m , \\
    &Q^{ab} = q\delta_{a\neq b} + Q \delta_{ab}, && \hat Q^{ab} = -\hat q\delta_{a\neq b} + \hat Q \delta_{ab} . 
\end{align}
and takes the limit $n\to 0$, reaching 
\begin{align}
    & \Psi_t = - \mathrm{Tr}(\hat m^{\top }m ) + \frac{1}{2}\mathrm{Tr}(\hat Q Q ) + \frac{1}{2} \mathrm{Tr}( \hat q^\top q),\\
    & \Psi_K = - \frac{1}{2} \log \left(\mathrm{det}(\mathbb{I}_F + \hat Q + \hat q ) \right) + \frac{1}{2} \mathrm{Tr} \left((\mathbb{I}_F+ \hat Q + \hat q)^{-1}(\hat m \hat m^\top +\hat q) \right),\\
    & \Psi_y = \mathbb{E}_{\theta^*, \xi}\left[\log \left(\mathbb{E}_\theta \left[ \mathrm{exp}\left( \theta^\top (m\theta^* + q^{\frac{1}{2}}\xi) - \frac{1}{2}\theta^\top q \theta \right)\right] \right) \right],
\end{align}
with $\xi\sim \mathcal{N}(0,\mathbb{I}_F)$. A main simplification, known as Nishimori condition, follows from the exchangeability, conditional on the training data, of the planted parameters and posterior replicas: we have $m=q, Q=\mathbb{I}_F, \hat m =\hat q$, also leading to $\hat Q = 0$. Therefore, the free entropy reads
\begin{align}\label{Eq:BO_free_entropy}
    \Phi = \underset{q,\hat q}{\mathrm{extr}} &\left[ -\frac{1}{2}\mathrm{Tr}(\hat q q) + \frac{1}{2}\mathrm{Tr}(\hat q) - \frac{1}{2} \log\left(\mathrm{det}(\mathbb{I}_F + \hat q)\right) \right. \notag \\
    & \left. + \alpha \mathbb{E}_{\theta^*, \xi }\left[\log\!\left(\mathbb{E}_\theta\!\left[\mathrm{exp}(\theta^\top (q\theta^* + q^\frac{1}{2}\xi )-\frac{1}{2}\theta^\top q \theta ) \right] \right) \right]\right].
\end{align}
Thus, for independent posterior samples $K^a \sim P(K\mid \mathcal{D})$, extremization of the free entropy (\ref{Eq:BO_free_entropy}) gives the summary statistics
\begin{align}
    \frac{K^a K^{*\top}}{D}=q , && \frac{K^a K^{b\top}}{D}= q + (\mathbb{I}_F - q)\delta_{ab}.
\end{align}
Given a new test sample X with, by definition,
\begin{align}
    X_\ell = \delta_{\ell\epsilon} \frac{K^{*\top}\theta^*}{\sqrt{D}} + z_\ell
\end{align}
where $z_\ell \sim \mathcal{N}(0,\mathbb{I}_D)$ and $\theta^* \sim P_\theta$, the law of the projections is determined by the summary statistics 
\begin{align}
    \frac{K^a X_\ell}{\sqrt{D}}\sim \delta_{\ell\epsilon}q\theta^* + q^{1/2}\xi_\ell + (\mathbb{I}_F-q)^{1/2} \eta_\ell^a 
\end{align}
with $\xi_\ell, \eta_\ell^a\sim\mathcal{N}(0,\mathbb{I}_F)$. Going back to the evaluation of the BO estimator, the posterior average in (\ref{Eq:BO_estimator}) can then be written 
\begin{align}
    \mathbb{E}_{K\mid \mathcal{D}}\mathbb{E}_\theta \left[e^{\theta^\top \frac{K X_\ell}{\sqrt{D}} - \frac{1}{2}\theta^\top \frac{KK^\top}{D}\theta} \right] &= \mathbb{E}_\theta \mathbb{E}_{\eta}\left[e^{ \theta^\top(\delta_{\ell\epsilon}q\theta^* + q^{1/2}\xi_\ell) + \theta^\top (\mathbb{I}_F-q)^{1/2}\eta   - \frac{1}{2}\theta^\top\theta }\right]\\
    & = \underbrace{\mathbb{E}_\theta \left[ e^{\theta^\top(\delta_{\ell\epsilon}q\theta^* + q^{1/2}\xi_\ell) - \frac{1}{2}\theta^\top q \theta  } \right]}_{p_\ell}.
\end{align}
Fixing the structured token $\epsilon=1$, giving $y=X_1$ without loss of generality, the BO error (\ref{Eq:BO_error}) can then be expressed as
\begin{align}
    \mathcal{E}_\mathrm{BO} = \mathbb{E}_{X, \mathcal{D}}\left[\frac{1}{2D}\left\lVert X_1 - \hat y_\mathrm{BO}(X, \mathcal{D})\right\rVert^2_2 \right] = \mathbb{E}_{\theta^*, \{\xi_\ell\}_{\ell=1}^L}\left[\frac{1}{2D}\left\lVert X_1 - \sum_{\ell=1}^L\frac{p_\ell}{\sum_{\ell'=1}^L p_{\ell'}}X_\ell\right\rVert_2^2 \right].
\end{align}
Expanding the square, using $X_\ell^\top X_{\ell'}/D\to \delta_{\ell\ell'}$ and $\mathbb{E}_{\theta^*,\{\xi_\ell\}}[\sum_\ell p_\ell^2/(\sum_jp_j)^2] =\mathbb{E}_{\theta^*,\{\xi_\ell\}}[p_1/\sum_jp_j]$, we reach
\begin{align}
    \mathcal{E}_\mathrm{BO} = \frac{1}{2} \mathbb{E}_{\theta^*, \{\xi_\ell\}_{\ell=1}^L}\left[1- \frac{p_1}{\sum_{\ell=1}^Lp_\ell}\right], && p_\ell = \mathbb{E}_\theta \left[ e^{\theta^\top(\delta_{\ell1}q\theta^* + q^{1/2}\xi_\ell) - \frac{1}{2}\theta^\top q \theta  } \right].
\end{align}

\subsubsection{Flipping spike model}
In the case of the flipping spike model where 
\begin{align}
    \theta = 
    \begin{pmatrix}
        \sqrt{\nu_1}\\ s\sqrt{\nu_2}
    \end{pmatrix},
    \theta^* = 
    \begin{pmatrix}
        \sqrt{\nu_1}\\ s^*\sqrt{\nu_2}
    \end{pmatrix},
    &&
    P(s=\pm 1) = P(s^*=\pm 1) = \frac{1}{2},
\end{align}
averaging over the random variables $s,s^*$, and using $\mathbb{E}_\xi \left[ (q^{1/2}\xi)_2^2\right]=q_{22}$, we reach 
\begin{align}
    \Phi = \underset{q,\hat q}{\mathrm{extr}} &\left[ -\frac{1}{2}\mathrm{Tr}(\hat q q) + \frac{1}{2}\mathrm{Tr}(\hat q) - \frac{1}{2} \log\left(\mathrm{det}(\mathbb{I}_F + \hat q)\right) \right. \notag \\
    & \left. + \alpha \frac{\nu_1 q_{11}- \nu_2 q_{22}}{2} + \alpha \mathbb{E}_{z}\left[\log\!\left(\cosh\!\left(\nu_2q_{22} + \sqrt{\nu_2 q_{22}} z \right) \right) \right]\right]
\end{align}
where we introduced $z\sim \mathcal{N}(0,1)$. The extremization of the free entropy then leads to the following self-consistent equations: 
\begin{align}\label{Eq:SE_BO}
    \begin{cases}
        q_{11} = \frac{\hat q_{11}}{1+\hat q_{11}}\\
        q_{22} = \frac{\hat q_{22}}{1+\hat q_{22}}
    \end{cases}
    &&
    \begin{cases}
        \hat q_{11} = \alpha \nu_1\\
        \hat q_{22} = \alpha \nu_2 \mathbb{E}_z\left[\tanh^2(\nu_2 q_{22}+ \sqrt{\nu_2 q_{22}}z) \right].
    \end{cases}
\end{align}
Similarly, expliciting the expression of $P_\theta$ gives the BO error for the flipping spike model:
\begin{align}\label{Eq:error_BO_flip}
    \mathcal{E}_\mathrm{BO} = \frac{1}{2} \mathbb{E}_{ \{z_\ell\}_{\ell=1}^L}\left[1- \frac{e^{\nu_1 q_{11}+\sqrt{\nu_1 q_{11}}z_{11}}\cosh(\nu_2q_{22}+\sqrt{\nu_2q_{22}}z_{12})}{\sum_{\ell=1}^L e^{\delta_{\ell1}\nu_1 q_{11}+\sqrt{\nu_1 q_{11}}z_{\ell1}}\cosh(\delta_{\ell1}\nu_2q_{22}+\sqrt{\nu_2q_{22}}z_{\ell2})}\right],
\end{align}
with $z_\ell \sim \mathcal{N}(0, \mathbb{I}_2)$. 
\begin{figure}[tbp]
    \centering
    \includegraphics[width=\linewidth]{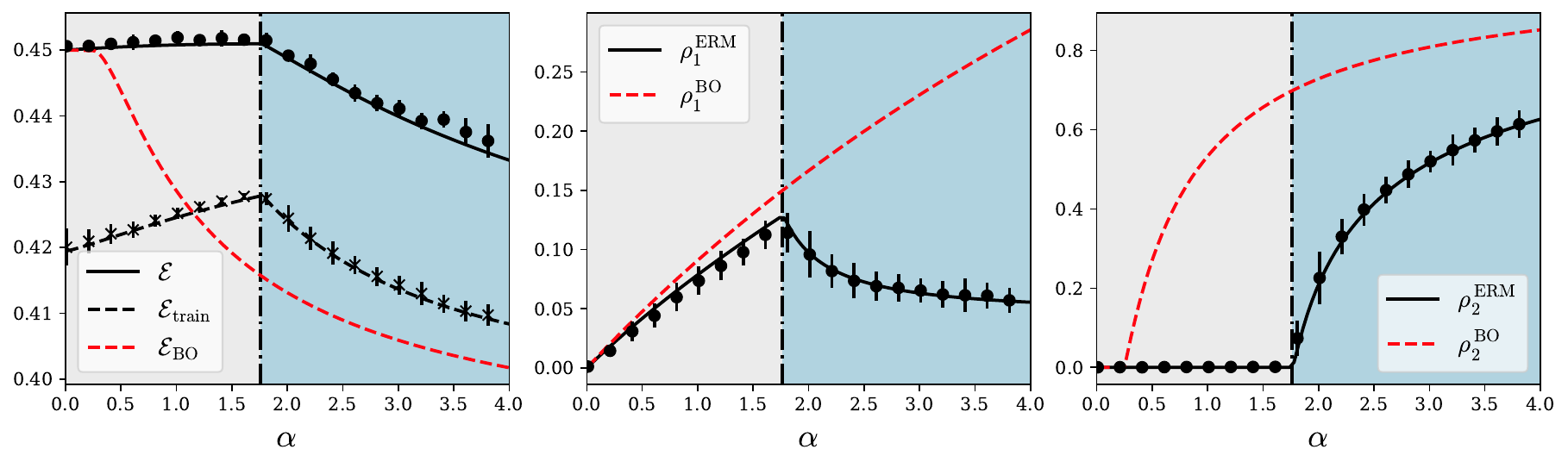}
    \caption{\textbf{Bayes-optimal estimator for the flipping spike model:} $\nu_1=0.1$, $\nu_2=2.0$, $L=10$, $H=2$, $\lambda=0.2$. Solid lines correspond to theoretical predictions (Res.~\ref{res:self-consistent_equations}, App. \ref{app:bayes_optimal}). Markers with errorbars correspond to training the model (Sec.~\ref{model}) with L-BFGS from random initialization in dimension $D=2000$ across ten seeds. \textbf{Left:} Test error $\mathcal E$ and training loss $\mathcal E_{\mathrm{train}}$ of the attention model, test error $\mathcal E_{\mathrm{BO}}$ of the Bayes-optimal estimator. \textbf{Center:} Normalized head-feature overlap $\rho_1^\mathrm{ERM}$ of the attention model and overlap $\rho_1^\mathrm{BO}$ of the Bayes-optimal estimator. \textbf{Right:} Normalized head-feature overlap $\rho_2^\mathrm{ERM}$ of the attention model and overlap $\rho_2^\mathrm{BO}$ of the Bayes-optimal estimator. Grey and blue identify the permutation-symmetric (\textcolor{PS}{PS}) and specialized (\textcolor{SPEC}{SPEC}) phases. Vertical lines mark the specialization threshold $\alpha_c^\mathrm{spec}$.}
    \label{Fig:A1}
\end{figure}
In Fig.~\ref{Fig:A1}, we compare the performances of the BO estimator in recovering the spikes, with $\rho_f^\mathrm{BO}=q_{ff}$ and $\mathcal{E}_\mathrm{BO}$ given by the solutions of (\ref{Eq:SE_BO}-\ref{Eq:error_BO_flip}), to the ones of minimizer of the empirical risk.  For comparison with the results of Sec.~\ref{Sec:asymptotic_characterization}, we quantify the recovery of planted features independently from weight norm in ERM with the normalized overlaps
\begin{align}\label{eq:rho_ERM}
    \rho_f^{\mathrm{ERM}} = \max_h \frac{(m_{hf}^\mathrm{ERM})^2}{q_{hh}^\mathrm{ERM}}
\end{align}
where we note $m^\mathrm{ERM}$ and $q^\mathrm{ERM}$ the summary statistics from Result \ref{res:self-consistent_equations}. 

\subsubsection{Non-isotropic Gaussian model}
In the case of the non-isotropic Gaussian model where 
\begin{align}
    \theta, \theta^* \sim \mathcal{N}(0,\Lambda), && \Lambda = \mathrm{diag}\left(\nu_1, ..., \nu_F \right),
\end{align}
averaging over the gaussian variables $\theta, \theta^*,\xi$, we reach 
\begin{align}
    \Phi = \underset{q,\hat q}{\mathrm{extr}} &\left[ -\frac{1}{2}\mathrm{Tr}(\hat q q) + \frac{1}{2}\mathrm{Tr}(\hat q) - \frac{1}{2} \log\left(\mathrm{det}(\mathbb{I}_F + \hat q)\right) + \frac{\alpha}{2}\mathrm{Tr}(\Lambda q) - \frac{\alpha}{2}\log\left(\det(\mathbb{I}_F + \Lambda q) \right)  \right].
\end{align}
The extremization of the free entropy then leads to the following self-consistent equations:
\begin{align}
    \begin{cases}
        q = \mathbb{I}_F - (\mathbb{I}_F + \hat q)^{-1}\\
        \hat q = \alpha \left[\Lambda - (\mathbb{I}_F + \Lambda q)^{-1} \Lambda \right],
    \end{cases}
\end{align}
with explicit solutions
\begin{align}\label{Eq:SE_BO_Gauss}
    q_{ff} = 
    \begin{cases}
        0, & \text{if }\; \alpha \nu_f^2 \leq 1,\\
        \frac{\alpha \nu_f^2- 1}{\nu_f (\alpha \nu_f + 1)}, & \text{if }\; \alpha \nu_f^2 > 1,
    \end{cases}
\end{align}
and $q_{hf}=0 \text{ for } h\neq f$. The BO error for the non-isotropic Gaussian model is
\begin{align}\label{Eq:error_BO_gauss}
    \mathcal{E}_\mathrm{BO} = \frac{1}{2} \mathbb{E}_{ \{z_\ell\}_{\ell=1}^L}\left[1- \frac{e^{\frac{1}{2}\sum_{f=1}^F\nu_f q_{ff}z_{1f}^2}}{e^{\frac{1}{2}\sum_{f=1}^F\nu_f q_{ff}z_{1f}^2}+ \sum_{\ell=2}^Le^{\frac{1}{2}\sum_{f=1}^F \frac{\nu_f q_{ff}}{1+\nu_f q_{ff}}z_{\ell f}^2}}\right],
\end{align}
with $z_\ell \sim \mathcal{N}(0, \mathbb{I}_F)$. 
\begin{figure}[tbp]
    \centering
    \includegraphics[width=\linewidth]{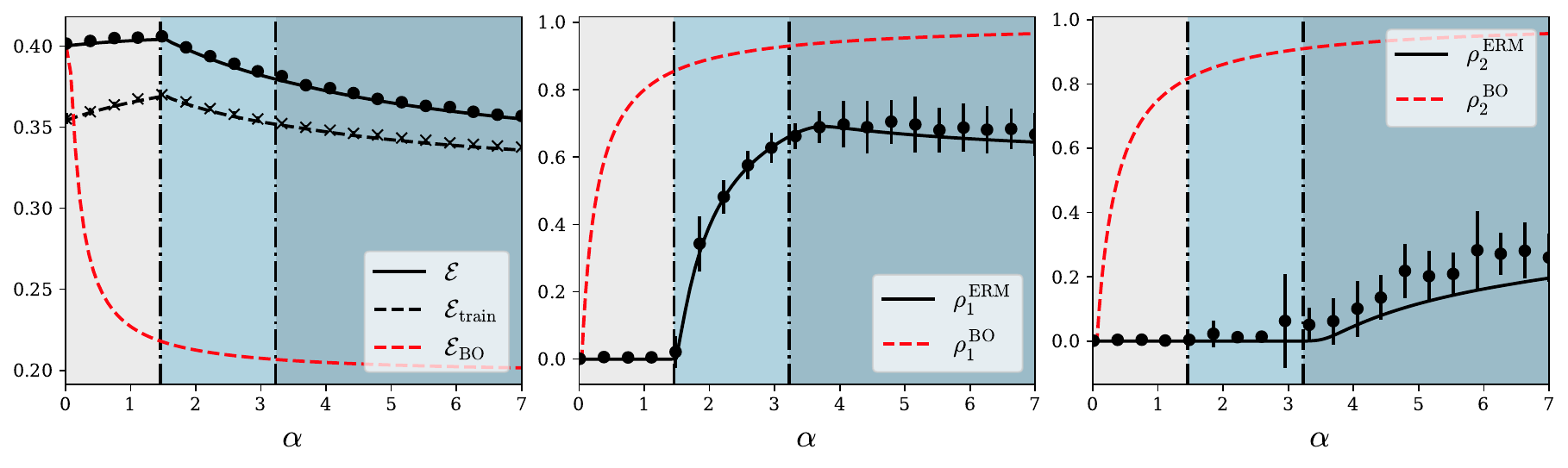}
    \caption{\textbf{Bayes-optimal estimator for the non-isotropic Gaussian model:} $\nu_1=5.0$, $\nu_2=4.0$, $L=5$, $H=4$, $\lambda=0.2$. Solid lines correspond to theoretical predictions (Res.~\ref{res:self-consistent_equations}, App. \ref{app:bayes_optimal}). Markers with errorbars correspond to training the model (Sec.~\ref{model}) with L-BFGS from random initialization in dimension $D=2000$ across ten seeds. \textbf{Left:} Test error $\mathcal E$ and training loss $\mathcal E_{\mathrm{train}}$ of the attention model, test error $\mathcal E_{\mathrm{BO}}$ of the Bayes-optimal estimator. \textbf{Center:} Normalized head-feature overlap $\rho_1^\mathrm{ERM}$ of the attention model and overlap $\rho_1^\mathrm{BO}$ of the Bayes-optimal estimator. \textbf{Right:} Normalized head-feature overlap $\rho_2^\mathrm{ERM}$ of the attention model and overlap $\rho_2^\mathrm{BO}$ of the Bayes-optimal estimator. Grey, light blue and darker blue identify the permutation-symmetric (\textcolor{PS}{PS}) and successive specialized (\textcolor{SPEC}{$1$-SPEC}, \textcolor{2-SPEC}{$2$-SPEC}) phases, respectively. Vertical dash-dotted lines mark the specialization thresholds $\alpha_c^{(1)}$ and $\alpha_c^{(2)}$.}
    \label{Fig:A2}
\end{figure}
In Fig.~\ref{Fig:A2}, we compare the performances of the BO estimator,  with $\rho_f^\mathrm{BO}=q_{ff}$ and $\mathcal{E}_\mathrm{BO}$ given by the solutions of (\ref{Eq:SE_BO_Gauss}-\ref{Eq:error_BO_gauss}), to the ones of minimizer of the empirical risk. For comparison with the results of Sec.~\ref{Sec:asymptotic_characterization}, we introduce the normalized overlaps
\begin{align}\label{eq:rho_ERM_gauss}
    \rho^\mathrm{ERM}_{f} = \max_h \frac{(m_{hf}^\mathrm{ERM})^2}{q_{hh}^\mathrm{ERM}}
\end{align}
where we note $m^\mathrm{ERM}$ and $q^\mathrm{ERM}$ the summary statistics from Result \ref{res:self-consistent_equations}.

\section{Model details and supplementary figures}
\label{App:Supplementary}
\subsection{Flipping spike model}
\begin{figure}[h]
    \centering
    \includegraphics[width=\linewidth]{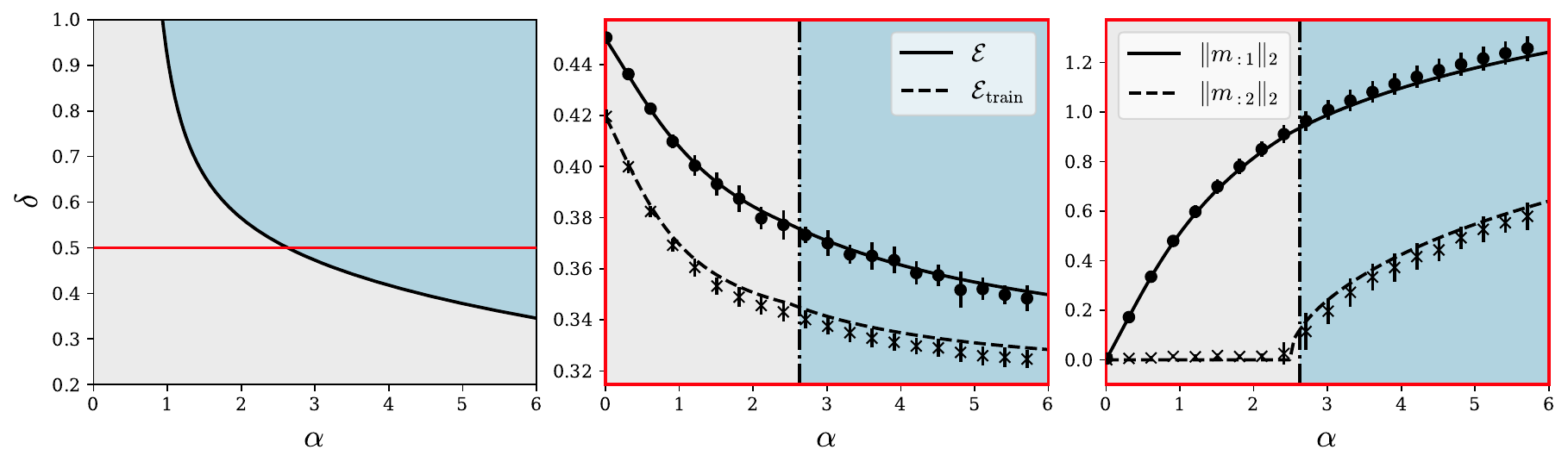}
    \caption{\textbf{Influence of signal composition on head specialization in the flipping spike model:} $\nu_{\mathrm{tot}}=4$, $L=10$, $H=2$, $\lambda=0.2$. Black solid lines correspond to theoretical predictions (Res~\ref{res:self-consistent_equations}). Markers with error bars correspond to training the model (Sec.~\ref{model}) with L-BFGS from random initialization in dimension $D=2000$ across ten seeds. \textbf{Left:} Phase diagram in sample complexity $\alpha$ and flipping ratio $\delta$. Phase boundaries are obtained by stability analysis of the different solutions (App.~\ref{App:Susceptibility}). The red horizontal line at $\delta=0.5$ corresponding to $\nu_1=\nu_2=2$, selects the cut shown in the other panels. \textbf{Center:} Test error $\mathcal E$ and training error $\mathcal E_{\mathrm{train}}$. \textbf{Right:} Norms of the head--feature overlaps $\lVert m_{:1}\rVert_2$ and $\lVert m_{:2}\rVert_2$. Vertical dash-dotted lines mark the specialization threshold $\alpha_c^\mathrm{spec}$ along the selected cut.}
    \label{Fig:B1}
\end{figure}

In the flipping spike model defined in Sec.~\ref{model}, a constant latent direction is learnable without head specialization while a flipping direction requires specialized heads to be used in the retrieval task. The signal strengths $(\nu_1,\nu_2)$ can be parametrized as
\begin{align}
    \nu_\mathrm{tot}=\nu_1+\nu_2, && \delta = \frac{\nu_2}{\nu_1+\nu_2},
\end{align}
to control total signal strength $\nu_\mathrm{tot}$ and flipping ratio $\delta$. More total signal facilitates the retrieval task, while larger flipping ratio increases the importance of specialization. For $\nu_2 = 0$, task requires no head specialization, and no phase transition occurs. For $\nu_1 = 0$, there is no constant direction and specialization is necessary for generalization, as in the non-isotropic Gaussian model. In Fig.~\ref{Fig:B1} we present the effect of the relative importance of the two features on specialization and performances.
\begin{figure}[!t]
    \centering
    \includegraphics[width=\linewidth]{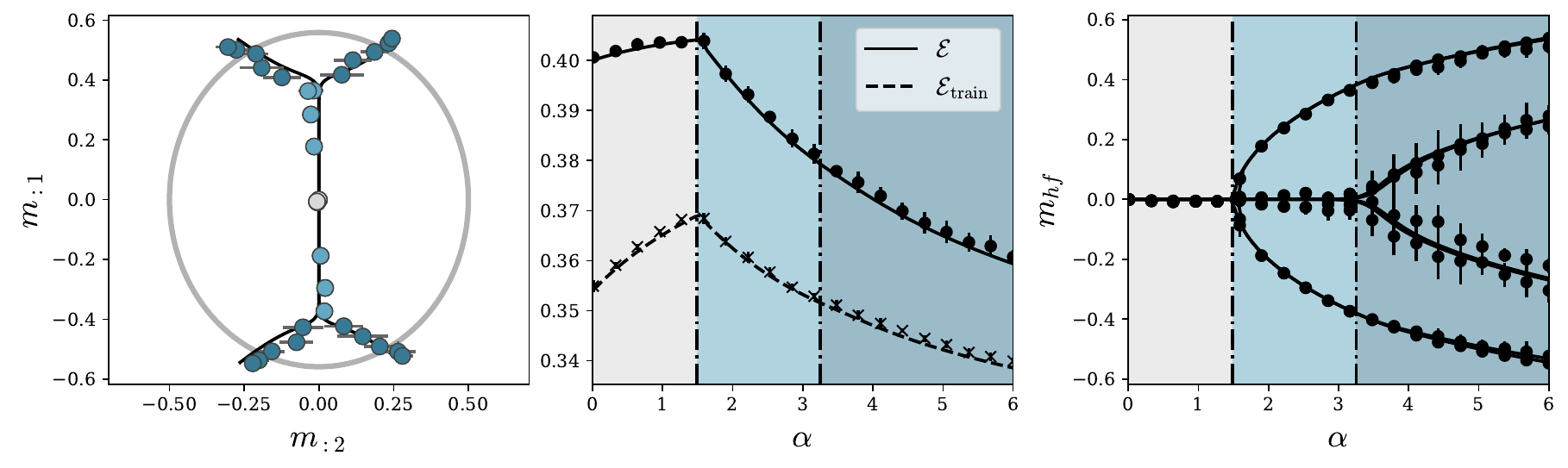}
    \caption{\textbf{Head-specialization phase transitions in the non-isotropic Gaussian model:} $\nu_1=5.0$, $\nu_2=4.0$, $L=5$, $H=4$, $\lambda=0.2$. Black solid lines correspond to theoretical predictions (Res.~\ref{res:self-consistent_equations}). Markers with error bars correspond to training the model (Sec.~\ref{model}) with L-BFGS from random initialization in dimension $D=4000$ across ten seeds. \textbf{Left:} Head-feature overlaps $(m_{:2},m_{:1})$ parametrized by sample complexity. Grey ellipse indicates the distribution anisotropy of possible draws of the planted spike $\hat{k}$. \textbf{Center:} Test error $\mathcal{E}$ and training error $\mathcal{E}_{\mathrm{train}}$. \textbf{Right:} Head-feature overlaps $m_{hf}$ for all heads and both latent features. Grey, light blue and darker blue identify the permutation-symmetric (\textcolor{PS}{PS}) and successive specialized (\textcolor{SPEC}{$1$-SPEC}, \textcolor{2-SPEC}{$2$-SPEC}) phases, respectively. Vertical dash-dotted lines mark the specialization thresholds $\alpha_c^{(1)}$ and $\alpha_c^{(2)}$.}
    \label{Fig:B2}
\end{figure}

\subsection{Non-isotropic Gaussian model}
\begin{figure}[!t]
    \centering
    \includegraphics[width=\linewidth]{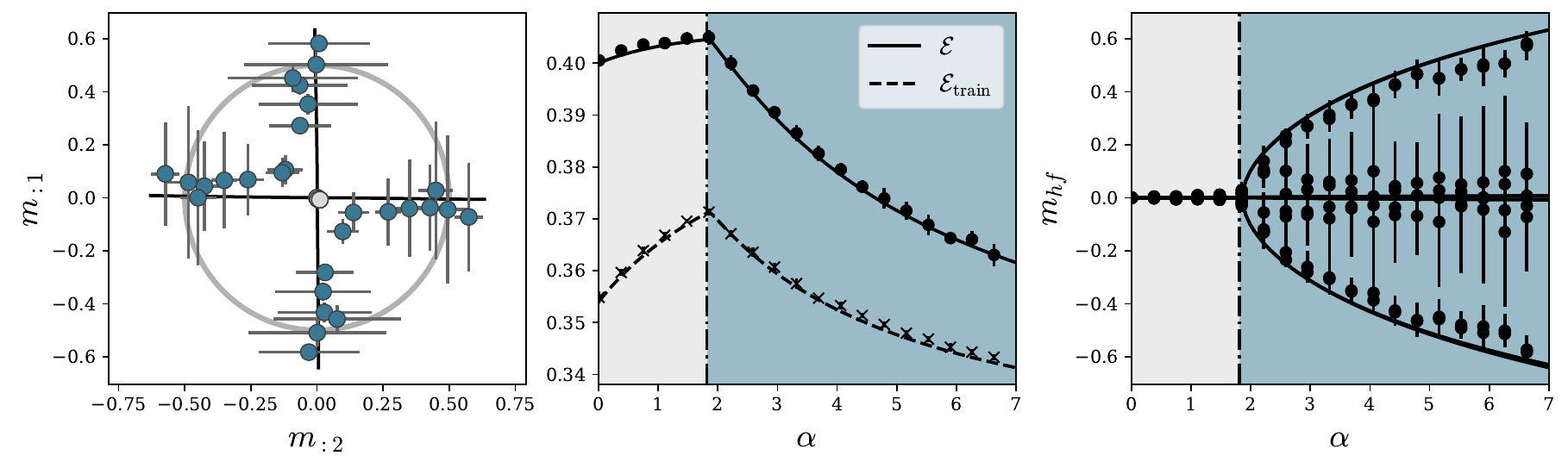}
    \caption{\textbf{Head-specialization phase transitions for isotropic Gaussian latent spikes:} $\nu_1=4.0$, $\nu_2=4.0$, $L=5$, $H=4$, $\lambda=0.2$. Black solid lines correspond to theoretical predictions (Res.~\ref{res:self-consistent_equations}). Markers with error bars correspond to training the model (Sec.~\ref{model}) with L-BFGS from random initialization in dimension $D=2000$ across ten seeds. \textbf{Left:} Head-feature overlaps $(m_{:2},m_{:1})$ parametrized by sample complexity. Grey circle indicates the distribution isotropy of possible draws of the planted spike $\hat{k}$. \textbf{Center:} Test error $\mathcal{E}$ and training error $\mathcal{E}_{\mathrm{train}}$. \textbf{Right:} Head-feature overlaps $m_{hf}$ for all heads and both latent features. Grey and darker blue identify the permutation-symmetric (\textcolor{PS}{PS}) and specialized (\textcolor{2-SPEC}{$2$-SPEC}) phases, respectively. Vertical dash-dotted lines mark the specialization thresholds $\alpha_c^{(1)}=\alpha_c^{(2)}$.}
    \label{Fig:B3}
\end{figure}

In Fig.~\ref{Fig:B2} we present additional figures for the non-isotropic Gaussian model, notably parametric curves for the head--feature overlaps.  In Fig.~\ref{Fig:B3} we present the same results for isotropic Gaussian distribution of the weights ($\nu_1=\nu_2$).

\section{Local stability and phase boundaries}
\label{App:Susceptibility}
We derive the local stability conditions for the solutions of the
self-consistent equations (\ref{eq:saddle_point_equations}) by tracking
small perturbations under the state-evolution (SE) map. These conditions
are used to compute the phase boundaries shown in
Figs.~\ref{Fig:2}, \ref{Fig:3}, and \ref{Fig:B1}.

\subsection{Flipping-spike model}

\paragraph{Stability against specialization}
Both PS and PSB satisfy $m_{:2}=0$, a condition preserved by the SE map. Denote an unperturbed solution by $(m_0,q_0,V_0)$, its local fields by $\Omega^0$, and its proximal by $Z'^0$, with $V_0=(\lambda\mathbb I_H+\hat V_0)^{-1}$. We perturb $m_{:2}=\delta m$ and denote the corresponding conjugate perturbation by $\delta\hat m$. By symmetry under $\theta_2\mapsto-\theta_2$, the remaining order parameters have no first-order variation. Writing $\theta_2=s\sqrt{\nu_2}$, with $s=\pm1$, the relevant linearized updates are
\begin{align}
    \begin{cases}
        \delta m=V_0\delta\hat m,\\[3pt]
        \displaystyle
        \delta\hat m
        =\frac{\alpha}{2}\sum_{s=\pm1}\mathbb E_\Xi
        \left[V_0^{-1}
        \left(\delta Z_1'^{(s)}-\delta\omega_1^{(s)}\right)
        s\sqrt{\nu_2}\right].
    \end{cases}
    \label{eq:stability_linear_updates}
\end{align}
We therefore need the first-order variations of the local fields and proximal. The fields are already linear in the perturbation:
\begin{align}
    \omega_\ell^{(s)}
    &=\omega_\ell^0+\delta_{\ell1}s\sqrt{\nu_2}\,\delta m,
    &\omega_\ell^0
    &=\delta_{\ell1}m^0_{:1}\sqrt{\nu_1}+q_0^{1/2}\xi_\ell.
\end{align}
Thus $\delta\omega_1^{(s)}=s\sqrt{\nu_2}\,\delta m$, while the other fields remain unchanged. To obtain $\delta Z'^{(s)}$, we expand the condition defining the proximal. Let $\ell(Z)$ denote the loss term in the Moreau minimization. Writing $Z'^{(s)}=Z'^0+\delta Z'^{(s)}$, its stationarity condition is
\begin{align}
    0={}&\nabla_Z\ell(Z'^0+\delta Z'^{(s)}) +(\mathbb I_L\otimes V_0^{-1})
    \left(Z'^0+\delta Z'^{(s)}-\Omega^0-\delta\Omega^{(s)}\right).
\end{align}
The zeroth-order terms cancel because $Z'^0$ satisfies the unperturbed stationarity condition. Taylor-expanding the gradient therefore gives, to first order,
\begin{align}
    \left[\nabla_Z^2\ell(Z'^0)+\mathbb I_L\otimes V_0^{-1}\right]
    \delta Z'^{(s)}
    =(\mathbb I_L\otimes V_0^{-1})\delta\Omega^{(s)}.
    \label{eq:stability_prox_linearization}
\end{align}
Since only $\omega_1$ is perturbed, we obtain
\begin{align}
    \delta Z_1'^{(s)}
    =(\left[\nabla_Z^2\ell(Z'^0)+\mathbb I_L\otimes V_0^{-1}\right]^{-1})_{11}V_0^{-1}\delta\omega_1^{(s)}
\end{align}
Therefore, at first order,
\begin{align}
    \delta\hat m
    &=\frac{\alpha}{2}\sum_{s=\pm1}\mathbb E_\Xi
    \left[
    V_0^{-1}\left((\left[\nabla_Z^2\ell(Z'^0)+\mathbb I_L\otimes V_0^{-1}\right]^{-1})_{11}V_0^{-1}-\mathbb I_H\right)
    \delta\omega_1^{(s)}s\sqrt{\nu_2}
    \right] \notag\\
    &=\alpha\nu_2V_0^{-1}\mathbb E_\Xi
    \left[(\left[\nabla_Z^2\ell(Z'^0)+\mathbb I_L\otimes V_0^{-1}\right]^{-1})_{11}V_0^{-1}-\mathbb I_H\right]\delta m.
\end{align}
Finally, using $\delta m=V_0\delta\hat m$, we get the SE map for the perturbation
\begin{align}
    \delta m^{t+1}=J_\mathrm{spec}(\alpha)\delta m^t,
    && J_\mathrm{spec}(\alpha)=\alpha \nu_2 \mathbb E_\Xi
    \left[(\left[\nabla_Z^2\ell(Z'^0)+\mathbb I_L\otimes V_0^{-1}\right]^{-1})_{11}V_0^{-1}-\mathbb I_H\right].
    \label{eq:flipping_stability}
\end{align}
The unspecialized solution is stable if the spectral radius of the linear operator $\delta m^t \to\delta m^{t+1} = J(\alpha)\delta m^t$ is less than one. Finding the critical sample complexity for the specialization transition then comes down to studying the largest eigenvalue of the matrix $J(\alpha)$ with the condition 
\begin{align}
    \lambda_{\mathrm{max}}[J_\mathrm{spec}(\alpha_c^\mathrm{spec})]=1,
\end{align}
where we noted $\lambda_\mathrm{max}[\cdot]$ the largest eigenvalue. Evaluating this condition on the PS or PSB solutions gives the PS--SPEC or PSB--SPEC stability respectively.

\paragraph{Stability against permutation symmetry breaking} At the PS solution with $m_{:2}=0$, we perturb towards PSB with  $q=q_0+\delta q\; e_-e_-^\top$, where $\delta q = q_\perp$ and $e_- = \frac{1}{\sqrt{2}}(1,-1)^\top$. Following similar steps as before leads to the SE map for the perturbation
\begin{align}
    &\delta q^{t+1} = J_\mathrm{psb}(\alpha) \delta q^t, \\ &J_\mathrm{psb}(\alpha)= 
    \alpha\mathbb E_\Xi \left[ \sum_{\ell,k=1}^L \left( e_-^\top\left[(\left[\nabla_Z^2\ell(Z'^0)+\mathbb I_L\otimes V_0^{-1}\right]^{-1})_{\ell k}V_0^{-1} -\delta_{\ell k}\mathbb I_H\right]e_- \right)^2 \right].
\end{align}
Again, the critical sample complexity is recovered through the condition 
\begin{align}
    \lambda_{\mathrm{max}}[J_\mathrm{psb}(\alpha_c^\mathrm{psb})]=1.
\end{align}

\subsection{Non-isotropic Gaussian model}
In the case of the non-isotropic Gaussian model, we study the stability against specialization to the next largest signal spike. Consider an $r$-SPEC solution $(m_r,q_r,V_r)$, for which $m_{r,:j}=0$ for all $j>r$. We perturb towards alignment with the next feature with $m_{:r+1}=\delta m_{:r+1}$. Since $\theta_{r+1}\sim\mathcal N(0,\nu_{r+1})$ is centered and independent of the recovered features, the remaining order parameters have no first-order variation. Denoting by $Z'_r$ the proximal evaluated at $(m_r,q_r,V_r)$; following similar steps as before leads to the SE map for the perturbation
\begin{align}
    &\delta m_{:r+1}^{t+1} = J_r(\alpha)\,\delta m_{:r+1}^{t},\\
    &J_r(\alpha) =\alpha\nu_{r+1} \,\mathbb E_{\theta, \Xi}\!\left[ \left( \left[\nabla_Z^2\ell(Z'_{r}) +\mathbb I_L\otimes V_r^{-1}\right]^{-1} \right)_{11}V_r^{-1} -\mathbb I_H \right].
\end{align}
the critical sample complexity is recovered through the condition
\begin{align}
    \lambda_{\max}\!\left[J_r(\alpha_c^{(r+1)})\right]=1.
\end{align}

\section{Link between statistical transitions and dynamical emergence}
\label{App:Dynamics}
We consider the case of the non-isotropic Gaussian model where $m=0, q=0$ is a saddle point of the loss, by symmetry $K\mapsto -K$. We focus on the first specialization and assume it happens only along the first feature $k^*_1$. We introduce the largest curvature $C\in\mathbb R^+$ of minus the population loss $\mathcal{E}_\mathrm{pop}$ in the direction of the first feature $k^*_1$, defined as
\begin{align}
C = \lambda_\mathrm{max}\left[-(k^*_1)^\top(\nabla^2_{K,K}\mathcal{E}_\mathrm{pop}(0))k^*_1\right]\;,\qquad \mathcal{E}_\mathrm{pop}(K) = \mathbb{E}_{X, y}\left[ \frac{1}{2D} \lVert y- f_{K}(X)\rVert_2^2\right] \label{eqApp:C}
\end{align}
where $\lambda_\mathrm{max}[\cdot]$ is the largest eigenvalue and $\nabla^2_{K,K}$ the Hessian.

\subsection{Online gradient flow}
\label{partApp:spéGF}
We consider the gradient-flow limit of unregularized online SGD. At time $\tau$ the update is
\begin{align}
    \frac{\mathrm dK}{\mathrm d\tau} = -D\nabla_K\mathcal{E}_\mathrm{pop}(K)\ , \label{eqApp:flot}
\end{align}
with the weights being independently initialized as $k_h(\tau=0)\sim\mathcal N(0,\eta^2\mathbb{I}_D)$ for any $h$ and for $\eta>0$. The normalization $D$ in eq.~\ref{eqApp:flot} insures that the time variation of the overlap $m$ are of order $\mathcal O_D(1)$.

We follow \citet{sagitova2026specializationsoftmaxattentionheads}. They show that the alignment $m$ follows the evolution
\begin{align}
\frac{\mathrm dm}{\mathrm d\tau} = -\nabla_m\mathcal{E}_\mathrm{pop}\;,\quad m_h(\tau=0)\sim\mathcal N(0,D^{-1}\eta^2\mathbb{I}_F)
\end{align}
and that in the limit $\eta\to 0$ the weights stay in the span of the features $k_h\in\mathrm{span}(k_f^*)_{f=1}^F$ with $q=mm^\top$. $m=0$ is an unstable fixed-point; we perform a linear stability analysis around it. We project $m$ onto the direction corresponding to the fastest descent direction, defined as the eigenvector of the maximal eigenvalue in eq.~\ref{eqApp:C}. We write $\bar m\in\mathbb R$ its projection. We have the following exponential escape
\begin{align}
\partial_\tau m(\tau) &= -\nabla_m\mathcal{E}_\mathrm{pop}(m(\tau)) \\
&= -\nabla_{m,m}^2\mathcal{E}_\mathrm{pop}(0)m(\tau) \\
&= -m(\tau)K^*\nabla_{K,K}^2\mathcal{E}_\mathrm{pop}(0)(K^*)^\top \\
\bar m(\tau) &= \bar m(0)e^{C\tau}
\end{align}
For any small enough threshold $\varepsilon>0$, independent of $\eta$, we define $\tau^\mathrm{spec}$ the specialization time to be the first $\tau$ such that $|\bar m|=\varepsilon$. We have
\begin{align}
\varepsilon &=|\bar m(0)|e^{C\tau^\mathrm{spec}} \\
\tau^\mathrm{spec} &= \frac{1}{C}\left(\log\varepsilon-\log|\bar m(0)|\right)
\end{align}
We define the rescaled specialization time $\tilde\tau^\mathrm{spec}$ as
\begin{align}
\tilde\tau^\mathrm{spec} &= \frac{\tau^\mathrm{spec}}{\log\sqrt D/\eta}
\end{align}
We have $|\bar m(0)|=\eta\bar{\bar m}/\sqrt D$ with $\bar{\bar m}$ a random variable independent of $\eta$ and $D$. In the limits $D\to\infty$ and $\eta\to 0$ we obtain that
\begin{align}
\tilde\tau^\mathrm{spec} &= \frac{1}{C}\left(\frac{\log\varepsilon}{\log\sqrt D/\eta}-\frac{\log\bar{\bar m}}{\log\sqrt D/\eta}+1\right) \to 1/C
\end{align}
that is to say the specialization time is only controlled by the Hessian of the loss at initialization. The random fluctuations due to the initialization and the effect of $\varepsilon$ vanish.

\subsection{ERM}
\label{partApp:spéERM}
We consider the static specialization in sample complexity. We consider the limit of large regularization $\lambda, \alpha\gg 1$ with $\alpha/\lambda=\Theta(1)$. In this limit, the model does not overfit data and $\mathcal E_\mathrm{train}=\mathcal{E}_\mathrm{pop}$. We also have $q=mm^\top$ and $D^{-1}||K||_F^2=\mathrm{Tr}q=||m||_F^2$, that is to say the weights $K$ are solely in the space of the ground truth $K^*$ and do not have components that fit noise.

We expand the training objective to the 2nd order in $m$.
\begin{align}
\frac{1}{D}\mathcal L_{\rm train} &= \alpha\mathcal{E}_\mathrm{pop}+\frac{\lambda}{2}||m||_F^2 \\
&= \alpha\mathcal{E}_\mathrm{pop}(0)+\frac{1}{2}\left(\alpha\nabla_{m,m}^2\mathcal{E}_\mathrm{pop}(0)+\lambda\mathbb{I}_H\otimes\mathbb{I}_F\right)\cdot (m,m) \\
&= \alpha\mathcal{E}_\mathrm{pop}(0)+\frac{1}{2}\left(\alpha K^*\nabla_{K,K}^2\mathcal{E}_\mathrm{pop}(0)(K^*)^\top+\lambda\mathbb{I}_H\otimes\mathbb{I}_F\right)\cdot (m,m) \label{eqApp:quadFormM}
\end{align}
We define $\alpha^\mathrm{spec}$ the smallest sampling ratio such that $m=0$ is no longer a local minimum of $\mathcal L_{\rm train}$. We only consider the specialization along $k_1^*$ and project $m$ onto $k_1^*$. The stability of the fixed-point $m=0$ is controlled by whether the smallest eigenvalue of the quadratic form eq.~\ref{eqApp:quadFormM} is negative. It reads $-\alpha C+\lambda$. This gives the critical sample complexity
\begin{align}
\alpha^\mathrm{spec} = \frac{\lambda}{C}
\end{align}
Last we rescale $\alpha$ as
\begin{align}
\tilde\alpha^\mathrm{spec} = \frac{\alpha^\mathrm{spec}}{\lambda}
\end{align}
and obtain that
\begin{align}
\tilde\alpha^\mathrm{spec} = \tilde\tau^\mathrm{spec} = 1/C.
\end{align}

\begin{figure}[h]
    \centering
    \includegraphics[width=0.75\linewidth]{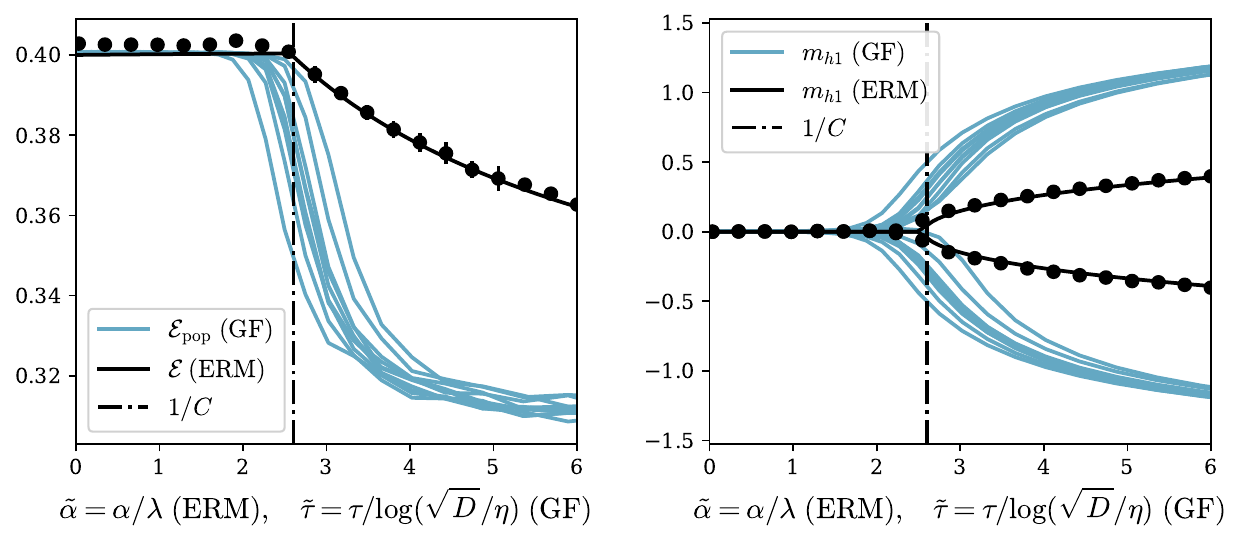}
    \caption{\textbf{Correspondence between $\alpha_c^\mathrm{spec}$ and $\tau_c^\mathrm{spec}$:} Population loss $\mathcal{E}_\mathrm{pop}$, test error $\mathcal{E}$ and head--feature overlap $ m_{h1}$ for online gradient flow as a function of rescaled time $\tilde \tau $ (ten independent runs) and for empirical risk minimizers as a function of rescaled sample complexity $\tilde \alpha$ (averaged over five instances). The inverse curvature $1/C$ is computed analytically in \citet{sagitova2026specializationsoftmaxattentionheads}. Gaussian model with $F=1$, $\nu_1=8$, $H=2$, $L=5$, $\lambda=3$, $D=5000$, $\eta=10^{-3}$.}
    \label{Fig:E1}
\end{figure}

The extension to more realistic scenarios, including online SGD and finite regularization $\lambda$, could be approached by a perturbation of this analysis; we expect the values of $\tilde \tau^\mathrm{spec}$ and $\tilde \alpha^\mathrm{spec}_c$ to be different but the underlying mechanism would stay the same. We provide a numerical comparison of the mapping between the ERM and the online gradient flow in Figure \ref{Fig:E1} at finite large $D$. We observe that specialization happens around the same $\tilde\tau^\mathrm{spec}$ and $\tilde\alpha^\mathrm{spec}$, whose values are well predicted by $1/C$. 

\section{Implementation details}
\label{App:Experiments}
\subsection{Experiments with solvable models}

\paragraph{Resolution of the self-consistent equations}
We obtain solutions for the set of self-consistent equations (\ref{eq:saddle_point_equations}) by iterating the associated state-evolution until convergence. The low-dimensional expectations over random variables $(\theta,\Xi)$ are evaluated with Monte-Carlo, with the proximal minimization performed numerically for each sample. Different solutions are first identified from uninformed initializations. Once their structure is known, branches can be followed using planted initialization, or by restricting the state-evolution updates on each solution. Their regularized training loss (or zero-temperature free entropy) identifies the global ERM minimum.

\paragraph{Numerical simulations}
For numerical simulations, we train the attention layer (\ref{eq:estimator}) on the regularized ERM objective (\ref{eq:ERM}), using synthetic datasets drawn from the model of Sec.~\ref{model}. For each realization, we sample the planted directions and generate $N=\lfloor\alpha D\rfloor$ training sequences, keeping both fixed throughout optimization. Training is implemented in PyTorch using full-batch L-BFGS with learning rate $0.5$, history size of $10$ steps, until convergence, starting from random initialization of the weights orthogonal to the planted directions. Weights are sampled from an isotropic Gaussian distribution and projected onto the orthogonal complement of the planted-feature span, removing finite-size initial alignments with the planted directions; the effect of this projection vanishes at large $D$. Final overlaps and training losses are computed directly, while test errors are evaluated on $1000$ separately generated sequences sharing the same latent directions. Each parameter setting is repeated over ten seeds, varying both the dataset and initialization. Before averaging empirical head--feature overlaps across seeds, we order symmetry-equivalent solutions consistently.

\subsection{Experiments with BERT MLMs}\label{App:BERT}

\paragraph{Fixed dataset experiment} We adapt the BERT masked-language-modeling experiments from \citet{ICLR2024_3e8df255}, which studied the emergence of syntactic attention structure (SAS) in training time, close to the online learning regime. To investigate an equivalent analysis in sample complexity, we train models on fixed datasets of $N$ sequences until the overfitting regime. To avoid data augmentation considerations, we also remove dynamic masking. Effective regularization is obtained through early stopping, evaluating the different metrics on checkpoint models with the lowest validation error along training trajectory. Repeating this procedure across dataset sizes and optimization seeds reveals the correspondence between emergence of specialized heads as a function of the number of samples and along training.

\paragraph{Training dataset} Training sequences are sampled from the preprocessed BookCorpus and English Wikipedia dataset of the reference experiment. Training datasets are nested subsets of $N$ sequences of $128$ tokens randomly sampled from the original $\sim9\mathrm{M}$ sequences dataset, with $2048$ disjoint validation sequences shared across runs. Masking selects $15\%$ of tokens, following the original $80/10/10$ replacement rule. 

\paragraph{Model and training} We train $\mathrm{BERT}_\mathrm{Small}$ from random initialization, with 4 layers, 8 heads per layer, hidden dimension 512, feed-forward dimension 2048, and dropout probability 0.1. Optimization uses AdamW with learning rate $10^{-4}$, weight decay $0.1$, and an effective batch size of $256$ sequences, obtained by accumulating four minibatches of $64$. The learning rate increases linearly over $1000$ warmup steps and then decreases linearly on a fixed $10^6$-step horizon. We train for $10000$ optimizer steps, saving and evaluating checkpoints every $250$ steps. Dataset sizes are $N\in\{1000, 3000, 6000, 9000, 12000, 15000, ..., 60000\}$ and $\mathrm{seeds}\in\{0,1,2,3,4\}$. Seeds vary initialization, minibatch order, and dropout. Given a sequence $\tilde X_\mu$ corrupted at positions $M_\mu$ from an original sequence $X_\mu$, the MLM loss is
\begin{align}\label{eq:MLM_loss}
    \mathcal{L}_\mathrm{MLM}(\Theta) = -\frac{1}{\sum_{\mu=1}^N\lvert M_\mu\rvert}\sum_{\mu=1}^N\sum_{i\in M_\mu}\log p_\Theta (x_{\mu i} \mid \tilde X_\mu),
\end{align}
where $p_\Theta(x_{\mu i} \mid \tilde X_\mu)$ is the vocabulary distribution at position $i$ predicted by the model with parameters $\Theta$. The training error $\mathcal{E}_\mathrm{train}$ therefore corresponds to the latest logged MLM loss in training, while the validation error $\mathcal{E}$ is the MLM loss computed on the validation dataset.

\paragraph{Syntactic attention structure} SAS is evaluated on a separate dataset of $1000$ fixed Wall Street Journal (WSJ) sentences from the Penn Treebank \citep{marcus1999treebank3}, with dependency parents and relation labels supplied in Stanford Dependencies format \citep{schuster-manning-2016-enhanced}. We follow the implementation from the repository of \citet{ICLR2024_3e8df255}. Token attention is first translated to word attention. Then, for head $h$ in layer $l$ and attention direction $d\in\{\rightarrow,\leftarrow\}$, the head-specific probe $\hat j_{h,l,d}$ for a target word $x_i$ is the other word with maximum attention in that direction:
\begin{align}
    \hat j_{h,l,d}(x_i)
    = \underset{j\neq i}{\arg\max}\,\alpha_{ij}^{h,l,d}.
\end{align}
Here $\alpha^{h,l}$ denotes the word-level attention matrix of head $h$ in layer $l$, with $\alpha_{ij}^{h,l,\rightarrow}=\alpha_{ij}^{h,l}$ and $\alpha_{ij}^{h,l,\leftarrow}=\alpha_{ji}^{h,l}$. Let $\mathcal R$ denote the set of syntactic relations present in the probe dataset, and $\mathcal D_R$ the collection of annotated child--parent pairs $(x_i,x_j)$ carrying relation $R$. The relation-specific UAS is
\begin{align}
    \mathrm{UAS}_R = \max_{h,l,d}\frac{1}{|\mathcal D_R|} \sum_{(x_i,x_j)\in\mathcal D_R} \mathbf{1}\{\hat j_{h,l,d}(x_i)=j\}.
    \label{eq:BERT_UAS_relation}
\end{align}
This measures how accurately best individual heads recover relation $R$. Different relations can select different heads, layers, and attention directions, allowing their specialization to be studied separately. The overall UAS selects the best head, layer, and attention direction across all dependencies:
\begin{align}
    \mathrm{UAS}
    = \max_{h,l,d}
      \frac{1}
      {\displaystyle\sum_{R\in\mathcal R}|\mathcal D_R|} \displaystyle\sum_{R\in\mathcal R}
      \sum_{(x_i,x_j)\in\mathcal D_R}
      \mathbf{1}\{\hat j_{h,l,d}(x_i)=j\}.
    \label{eq:BERT_UAS}
\end{align}
This measures the fraction of correctly recovered parents using a single head, layer, and attention direction. For relation-specific scores, relation labels select the probes, while correctness depends only on predicting the parent. These scores serve as benchmark proxies relating to head--feature overlaps; quantifying alignment between attention structure and syntactic relations structure, a structural latent feature of language datasets.

\begin{figure}[h]
    \centering
    \includegraphics[width=0.75\linewidth]{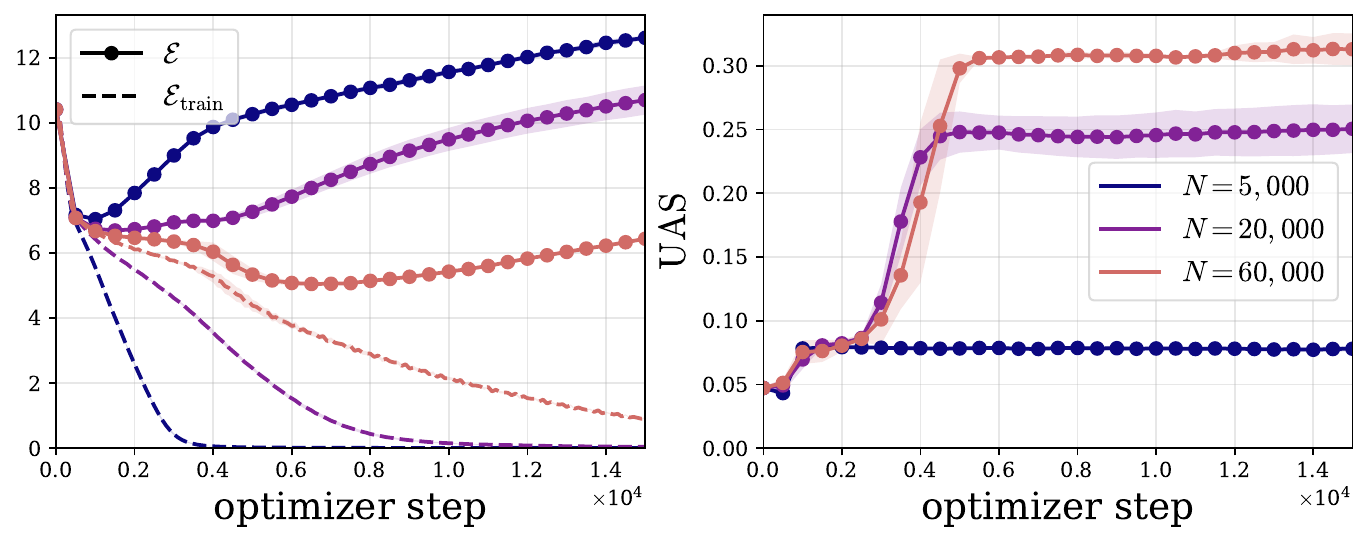}
    \caption{\textbf{Training dynamics on fixed datasets:} Separate runs of $\mathrm{BERT}_{\mathrm{Small}}$ models trained on datasets of $N\in\{5000,20000,60000\}$ sequences, following the training procedure described in App.~\ref{App:BERT} with $15000$ optimization steps and checkpoints every $500$ steps. \textbf{Left:} Validation MLM loss $\mathcal E$ at registered checkpoints and logged training loss $\mathcal E_{\mathrm{train}}$ along training. \textbf{Right:} Evolution of syntactic attention structure, measured by UAS. Larger datasets allow an abrupt emergence of SAS during training. Curves show averages across optimization seeds, with shaded bands indicating one standard deviation.}
    \label{Fig:F1}
\end{figure}

In Fig.~\ref{Fig:F1} we present the training dynamics for $\mathrm{BERT}_\mathrm{Small}$ at different dataset sizes. We observe the presence of drops in the error corresponding to the acquisition of SAS only at large enough dataset sizes while models trained on small datasets overfit the data early. As UAS plateaus, specialization remains until convergence. 

\begin{figure}[h]
    \centering
    \includegraphics[width=0.7\linewidth]{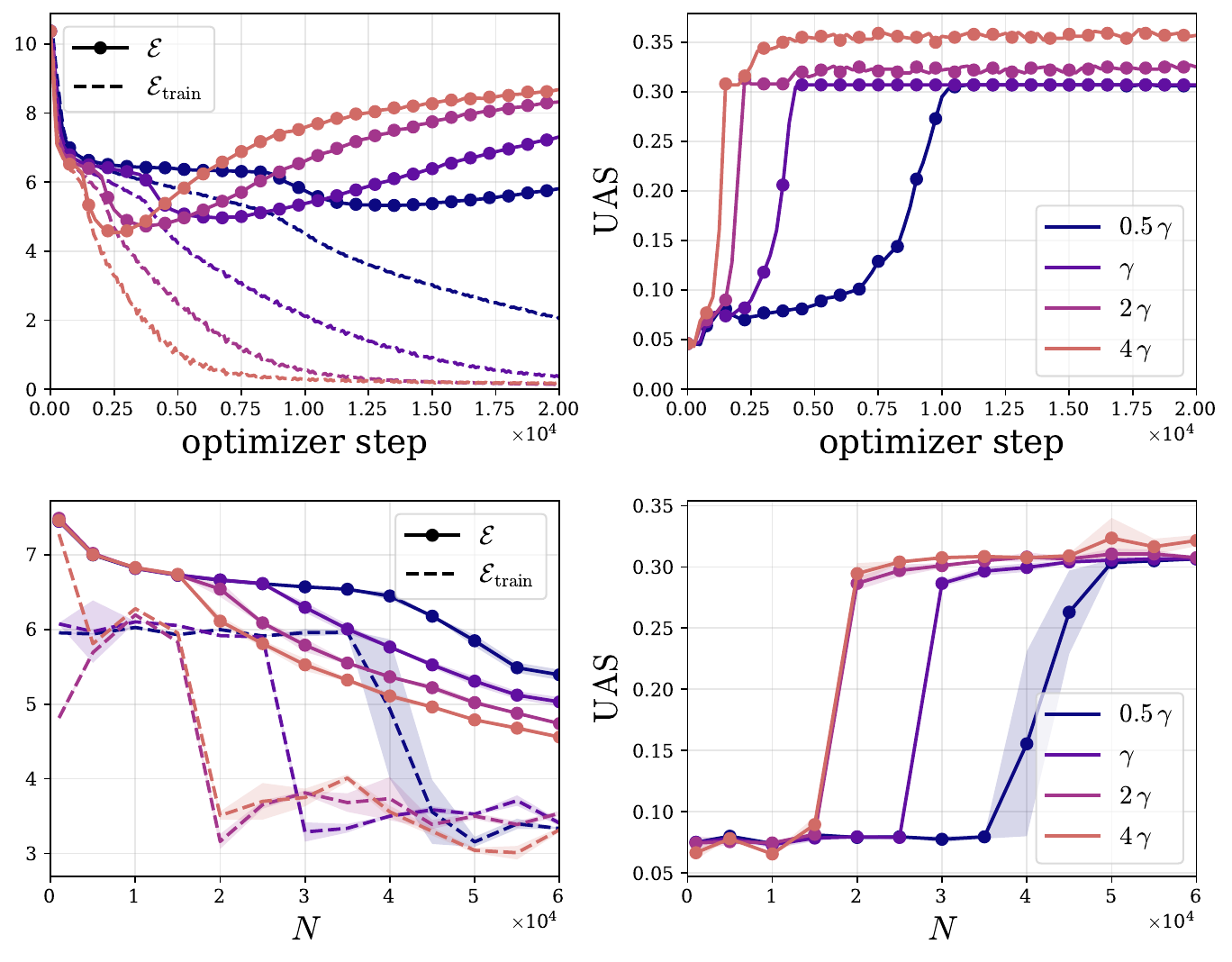}
    \caption{\textbf{Influence of the learning rate on specialization:} $\mathrm{BERT}_{\mathrm{Small}}$ models trained following the procedure described in App.~\ref{App:BERT}, with learning rates $\{0.5\gamma,\gamma,2\gamma,4\gamma\}$, where $\gamma=10^{-4}$. \textbf{Top:} Training dynamics on a fixed dataset of $N=60000$ sequences. \textbf{Bottom:} Models trained on fixed datasets of increasing size $N$. \textbf{Left:} Validation MLM loss $\mathcal E$ at registered checkpoints and logged training loss $\mathcal E_{\mathrm{train}}$. \textbf{Right:} Evolution of syntactic attention structure, measured by UAS. Curves show averages across two optimization seeds, with shaded bands indicating one standard deviation.}
    \label{Fig:F3}
\end{figure}

In Fig.~\ref{Fig:F3} we present the results of training $\mathrm{BERT}_\mathrm{Small}$ with varying learning rates. Head specialization is robust to the optimization dynamics: we observe emergence and a plateau in UAS, independently from the selected early-stopping time. As mentioned in Sec.\ref{discussion}, early stopping regularizes within the class of optimizers reached during a given optimization dynamics, and introduces a dependency on the learning rate in the critical number of samples for specialization. A further test of the sample-complexity interpretation would explore varying training procedures with a regularized objective and training controlled to convergence.

\begin{figure}[h]
    \centering
    \includegraphics[width=0.7\linewidth]{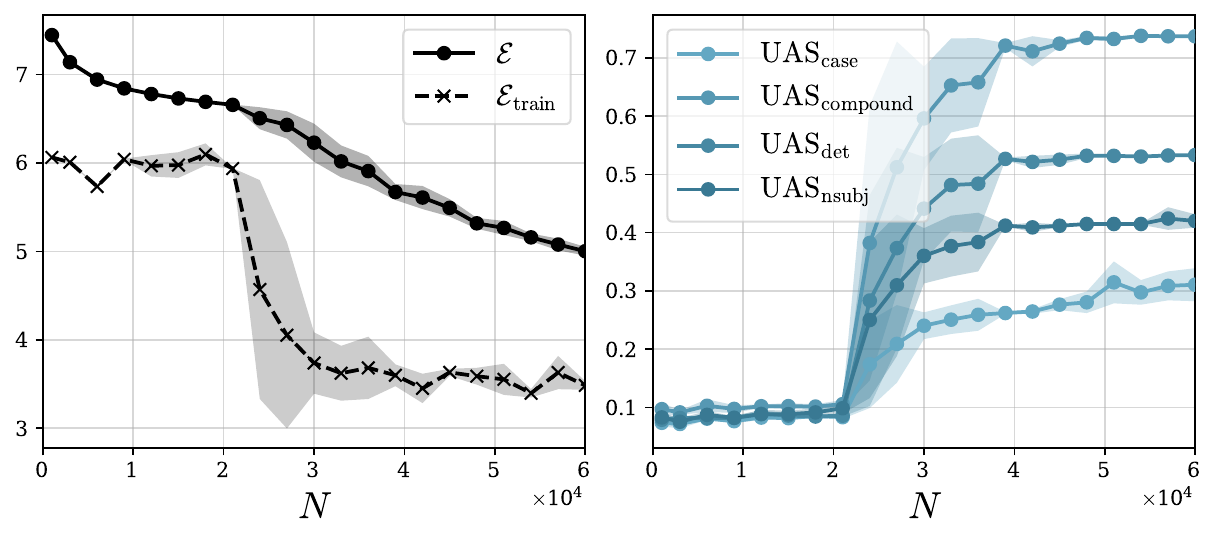}
    \caption{\textbf{Acquisition of different syntactic relations with increasing dataset size:} $\mathrm{BERT}_{\mathrm{Small}}$ models are trained on fixed datasets of increasing size $N$ following the training procedure described in App.~\ref{App:BERT}. \textbf{Left:} Validation MLM loss $\mathcal E$ at registered checkpoints and logged training loss $\mathcal E_{\mathrm{train}}$. \textbf{Right:} Relation-specific attachment scores for the four most frequent relations in the WSJ dataset: $\mathrm{case}$, $\mathrm{compound}$, $\mathrm{det}$, and $\mathrm{nsubj}$. Validation loss and attachment scores are evaluated with early stopping. Curves show seed averages, with shaded bands indicating one standard deviation.}
    \label{Fig:F2}
\end{figure}

Fig.~\ref{Fig:F2} details relation-refined $\mathrm{UAS}_R$ for the statistically most present relations in the dataset, exhibiting a same critical dataset size for emergence of SAS suggesting a coordinated acquisition of syntactic structure. It would be interesting to explore if different capabilities, maybe different types of syntactic rules, are obtained at different sample complexities.


\clearpage
\subsection{Experiments with sparse attention patterns}
\begin{figure}[h]
    \centering
    \includegraphics[width=0.7\linewidth]{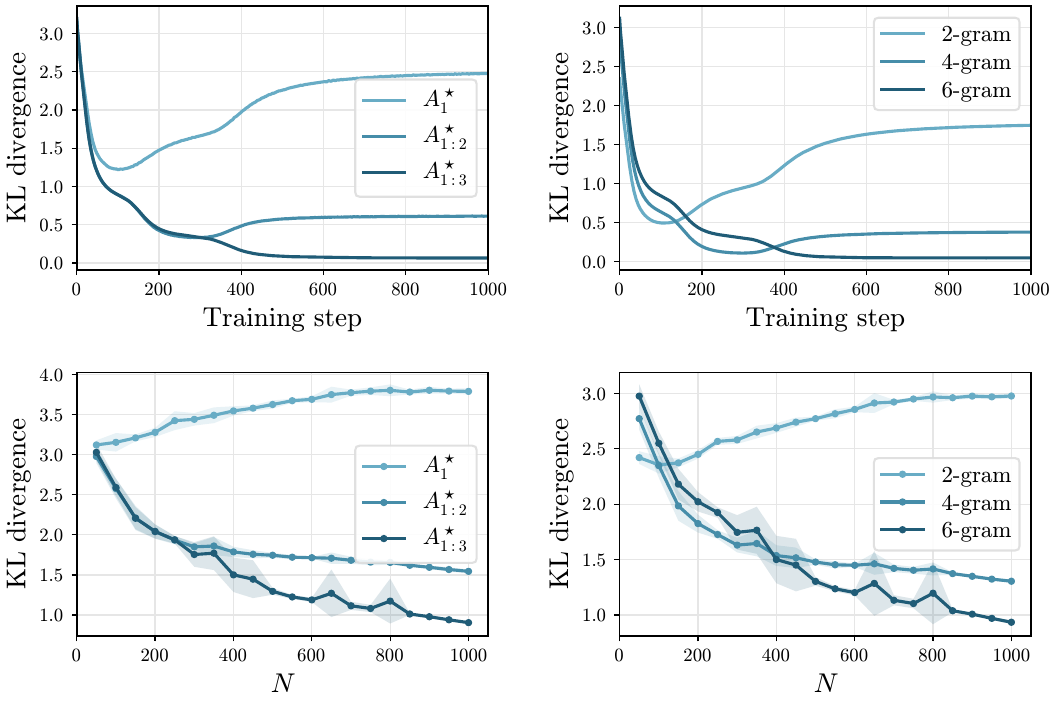}
    \caption{\textbf{Sequential specialization in learning of sparse attention patterns:} Next-token prediction on the synthetic Markov-chain task of \citet{yüksel2026incrementallearningsparseattention} (see their Fig.~3 for details). \textbf{Top:} Training dynamics of the full transformer trained on $N=9000$ sequences. \textbf{Bottom:} Transformer trained until convergence on fixed datasets of increasing size $N$, with $\ell_2$ regularization $\lambda=10^{-6}$. \textbf{Left:} KL divergence $\mathrm{KL}(f_{A^\star_{1:r}}\Vert f_{\mathrm{student}})$ with the ground-truth predictors $A^\star_{1:r}$ using the first $r=1,2,3$ positional groups. \textbf{Right:} KL divergence $\mathrm{KL}(f_{c\text{-gram}}\Vert f_{\mathrm{student}})$ where $c$-gram predictors are independently trained transformers restricted to the last $c=2,4,6$ tokens. Bottom curves average five seeds, with shaded bands indicating one standard deviation.}
    \label{Fig:E3}
\end{figure}

We reproduce the experiment of \citet{yüksel2026incrementallearningsparseattention} studying next-token prediction on a synthetic high-order Markov chain, where the prediction depends on three groups of past positions, ordered by decreasing statistical importance. Following their Figure 3, we measure the KL divergence between the student predictor $f_{\mathrm{student}}$ and ground-truth predictors $f_{A^\star_{1:r}}$ retaining only the first $r$ groups, and between the student and independently trained transformers $f_{c\text{-gram}}$ with restricted context length of $c$ tokens. We extend their online learning setting to fixed dataset sizes introducing a small $\ell_2$ regularization to stabilize convergence. As the dataset size increases, the converged predictors approach reference predictors using progressively more positional groups. In Fig.\ref{Fig:E3}, we reproduce the experiment as a function of the number of samples, and establish that specialization cascade also occurs in this setting. This tests the successive specialization phase-transitions and simplicity-bias predictions with correlated tokens and autoregressive loss.

\end{document}